\pdfoutput=1

\documentclass[11pt]{article}

\usepackage[final]{acl}

\usepackage{times}
\usepackage{latexsym}

\usepackage[T1]{fontenc}

\usepackage[utf8]{inputenc}

\usepackage{microtype}

\usepackage{inconsolata}

\usepackage{graphicx}
\usepackage{caption}
\usepackage{subcaption}

\usepackage{algorithm}
\usepackage{algorithmic}

\usepackage{newfloat}
\usepackage{listings}
\DeclareCaptionStyle{ruled}{labelfont=normalfont,labelsep=colon,strut=off} 
\floatstyle{ruled}
\newfloat{listing}{tb}{lst}{}
\floatname{listing}{Algorithm}

\usepackage{booktabs}
\usepackage{multirow}
\usepackage{makecell}
\usepackage{tcolorbox}
\tcbuselibrary{breakable, skins, theorems}
\usepackage{adjustbox}
\usepackage[framemethod=tikz]{mdframed}
\newenvironment{maintheorem}[1][]{%
  \begin{tcolorbox}[
    colback = blue!5,
    colframe = white,
    fonttitle = \bfseries,
    breakable = true]
  \begin{theorem}[#1]%
}{%
  \end{theorem}%
  \end{tcolorbox}
  \vspace{5pt}
}
\newenvironment{maintheoreminformal}[1][]{%
  \begin{mdframed}[backgroundcolor=blue!5, linecolor=blue!75!black, linewidth=2pt, roundcorner=5pt, skipabove=5pt]%
  \begin{theorem*}%
}{%
  \end{theorem*}%
  \end{mdframed}%
  \vspace{5pt}
}
\newenvironment{mainlemma}[1][]{%
  \begin{tcolorbox}[
    colback = gray!5,
    colframe = white,
    fonttitle = \bfseries,
    breakable = true]
  \begin{lemma}[#1]%
}{%
  \end{lemma}%
  \end{tcolorbox}
  \vspace{5pt}
}

\newenvironment{mainobservation}[1][]{%
  \begin{tcolorbox}[
    colback = green!5,
    colframe = white,
    fonttitle = \bfseries,
    breakable = true]
  \begin{observation}[#1]%
}{%
  \end{observation}%
  \end{tcolorbox}
  \vspace{5pt}
}
\newenvironment{maincoro}[1][]{%
  \begin{tcolorbox}[
    colback = red!5,
    colframe = white,
    fonttitle = \bfseries,
    breakable = true]
  \begin{corollary}[#1]%
}{%
  \end{corollary}%
  \end{tcolorbox}
  \vspace{5pt}
}
\newenvironment{mainassumption}[1][]{%
  \begin{tcolorbox}[
    colback = yellow!5,
    colframe = white,
    fonttitle = \bfseries,
    breakable = true]
  \begin{assumption}[#1]%
}{%
  \end{assumption}%
  \end{tcolorbox}
  \vspace{5pt}
}
\usepackage{framed}

\usepackage{dsfont}
\usepackage{amsfonts}
\usepackage{bbm}
\usepackage{amsmath}
\usepackage{amsthm}
\usepackage{mathtools}
\usepackage{enumitem}
\newcommand{\BOS}{\textsc{BOS}}
\newcommand{\EOS}{\textsc{EOS}}

 \newcommand{\Umax}{{U}_\mathrm{{max}}}
\newcommand{\Phuman}{P_\mathrm{{human}}}
\newcommand{\Pmodel}{P_\mathrm{{model}}} 
\newcommand{\human}{y^*}
\newcommand{\hmodel}{y^{\mathrm{m}}}
\newcommand{\hmonte}{\hat{y}^m}

\DeclareMathOperator*{\argmax}{arg\,max}
\newcommand{\ba}{\boldsymbol{\alpha}}
\newcommand{\bv}{\boldsymbol{v}}
\newcommand{\E}{\mathbb{E}}

\newcommand{\hm}{\hat{y}^m_{\mathrm{MAP}}} 
\newcommand{\maphuman}{y^*_{\mathrm{MAP}}}
\newcommand{\mapmodel}{y^m_{\mathrm{MAP}}}

\newcommand{\Yn}{\mathcal{Y}^\mathrm{n}_{\mathrm{ref}}}
\theoremstyle{plain}
\newtheorem{theorem}{Theorem}
\theoremstyle{theorem*}
\newtheorem*{theorem*}{Theorem}

\newtheorem{lemma}{Lemma}
\newtheorem{corollary}{Corollary}
\theoremstyle{definition}
\newtheorem{definition}{Definition}
\theoremstyle{assumption}
\newtheorem{assumption}{Assumption}
\theoremstyle{remark}

\theoremstyle{observation}
\newtheorem{observation}{Observation}

\usepackage[textsize=tiny]{todonotes}

\title{Theoretical Guarantees for Minimum Bayes Risk Decoding}
\author{
Yuki Ichihara$^{1}$\qquad Yuu Jinnai$^{2}$\qquad Kaito Ariu$^{2}$\qquad Tetsuro Morimura$^{2}$\qquad Eiji Uchibe$^{3}$\\
$^1$Nara Institute of Science and Technology\qquad $^2$CyberAgent\\\qquad$^3$Advanced Telecommunications Research Institute International\\
\texttt{ichihara.yuki.iu1@is.naist.jp}\\ \texttt{\{jinnai\_yu,kaito\_ariu,morimura\_tetsuro\}@cyberagent.co.jp}\\ \texttt{uchibe@atr.jp}
}

\begin{document}
\maketitle
\begin{abstract}
Minimum Bayes Risk (MBR) decoding optimizes output selection by maximizing the expected utility value of an underlying human distribution.
While prior work has shown the effectiveness of MBR decoding through empirical evaluation, few studies have analytically investigated why the method is effective.
As a result of our analysis, we show that, given the size $n$ of the reference hypothesis set used in computation, MBR decoding approaches the optimal solution with high probability at a rate of $\smash{O}\left(n^{-\frac{1}{2}}\right)$, under certain assumptions, even though the language space $\mathcal{Y}$ is significantly larger $|\mathcal{Y}|\gg n$.  
This result helps to theoretically explain the strong performance observed in several prior empirical studies on MBR decoding.
In addition, we provide the performance gap for maximum-a-posteriori (MAP) decoding and compare it to MBR decoding. The result of this paper indicates that MBR decoding tends to converge to the optimal solution faster than MAP decoding in several cases.
\end{abstract}
\section{Introduction}
Minimum Bayes Risk (MBR) decoding \cite{kumar-byrne-2002-minimum,kumar-byrne-2004-minimum} is a decision rule used to generate sequences from autoregressive probability models (e.g., LLMs). MBR decoding has been shown to produce high-quality texts in various directed text generation tasks, such as machine translation \cite{tromble-etal-2008-lattice, de-gispert-etal-2009-minimum, stahlberg-etal-2017-neural}, text summarization \cite{rush-etal-2015-neural, narayan-etal-2018-dont}, text simplification \cite{heineman-etal-2024-improving}, image captioning \cite{borgeaud-emerson-2020-leveraging}, instruction-following \cite{wu2025better}, and many of the systems submitted to the WMT competition\footnote{\url{https://machinetranslate.org/wmt}} adopt MBR decoding.  Numerous experiments have reported the advantages of MBR decoding over maximum-a-posteriori (MAP) decoding (e.g., beam search) \cite{ehling-etal-2007-minimum,  eikema-aziz-2020-map, muller-sennrich-2021-understanding, eikema-aziz-2022-sampling, bertsch-etal-2023-mbr}.

Experimental results confirm that the larger the number of candidates and hypothesis sets collected, the better performance \cite{eikema-aziz-2022-sampling,freitag-etal-2022-high}.
However, there is no theoretical explanation for the convergence rate of approaching optimal output.
The answers to this question are the number of elements in the candidate and the hypothesis set in this paper.
Our results show the following theorem.
\begin{maintheoreminformal}(Convergence Rate of MBR Decoding; Informal)
    Under certain assumptions, MBR decoding approaches the optimal solution with high probability at a rate of $\smash{O}\left(n^{-\frac{1}{2}}\right)$ for the size $n$ of the reference hypothesis set.
\end{maintheoreminformal}
This theoretical result is consistent with the empirical results of previous studies \cite{eikema-aziz-2022-sampling,freitag-etal-2022-high}.
We also confirm that if the human distribution is similar to the model distribution, the performance of MBR decoding can be improved, as indicated by \citet{ohashi-etal-2024-true}. 
In addition, we derive the convergence rate of the optimal output for MAP decoding and compare it to MBR decoding. Our results show that MBR decoding tends to converge faster than MAP decoding in several cases.

Specifically, our main contributions are that we provide high probability and expected regret's upper bounds by MBR decoding in several cases (Theorem~\ref{theorem:bound3}, Theorem~\ref{theorem:bound}, and Corollary ~\ref{propotion:mbr}) and we compare the performance gap and convergence rate of MBR decoding and MAP decoding within the same framework of the upper bound we derived in Section~\ref{section:map_mbr}.

In summary, there are few theoretical analyses of MBR decoding, and thus a comprehensive theoretical framework has yet to be fully established. 
Through these contributions, we believe that this study offers new perspectives that advance the understanding of MBR decoding.

Moreover, our theoretical results on MBR decoding have broad practical relevance. For example, even if advanced models such as GPT-8 emerge or if we encounter challenging tasks like ancient Japanese language generation, our work demonstrates that MBR decoding remains a viable option. This finding is practically beneficial, as it confirms that MBR decoding is not only theoretically grounded with convergence rate guarantees but also adaptable to a wide range of applications.

\section{Background and Notations}
Text generation involves producing an output sequence based on an input sequence, the set of input sequences is defined by $\mathcal{X}$. Probabilistic text generators define a probability distribution over the output space of hypotheses $\mathcal{Y}$. The set of complete hypotheses $\mathcal{Y}$ is:
\begin{equation*}
    \mathcal{Y} := \{\BOS \circ \bv \circ \EOS | \bv \in \mathcal{V}^*\}.
\end{equation*}
where $\circ$ is a string concatenation and $\mathcal{V}^*$ is the Kleene closure of a set of vocabulary $\mathcal{V}$. 
The goal of decoding is to find the best hypothesis for a given input. 
For simplicity 
we write $\mathcal{M}^{\mathcal{X}}_{\mathcal{Y}}$ to denote a set of conditional probability distributions over a finite set $\mathcal{Y}$, given $\mathcal{X}$ as context sets.
and $\smash{O}(n)$ is Big O notation.

\subsection{MBR Decoding}
Let $\mathcal{X}$ denote the input space and $\mathcal{Y}$ the output space. Given an input $x \in \mathcal{X}$, a probabilistic model defines a distribution $p\in\mathcal{M}^{\mathcal{X}}_{\mathcal{Y}}$ over possible outputs $y' \in \mathcal{Y}$. The goal of Bayes Risk minimization in structured prediction and sequence generation tasks is to select an output that minimizes the expected loss relative to the true distribution \cite{bach2024learning}.

For a loss function $\ell: \mathcal{Y} \times \mathcal{Y} \to \mathbb{R}$, the Bayes Risk is defined as:
\begin{equation*}
\mathcal{R}(y \mid x) = \E_p\left[\ell(y', y) \right].
\end{equation*}
\begin{equation*}
y^* = \arg\min_{y \in \mathcal{Y}} \mathcal{R}(y \mid x)
\end{equation*}
If the goal is to maximize some utility function $u$ rather than to minimize a loss, it can also be interpreted as a performance metric $\ell = -u $.

The objective of MBR decoding is similar to the Bayes Risk, finding the output that maximizes the expected utility, thereby effectively minimizing risk \cite{kumar-byrne-2002-minimum,kumar-byrne-2004-minimum}.

The procedure consists of two key components: the human distribution $\Phuman \in \mathcal{M}^{\mathcal{X}}_\mathcal{Y}$ given input $x\in \mathcal{X}$ and a utility function. For simplicity, let $P(y \mid x) = P(y)$, since $x$ is fixed in this paper.
The utility function evaluates the quality of a candidate output $\mathcal{H} \subseteq \mathcal{Y}$ with respect to a reference output $\mathcal{Y}$.  
In this paper, we assume that the candidate hypotheses $\mathcal{H}$ are identical to the reference outputs $\mathcal{Y}$.  
Ideally, MBR decoding selects the optimal hypothesis by maximizing its expected utility over the distribution of human references:
\begin{align}
    u_h(y) &= \sum_{y' \in \mathcal{Y}} u(y, y')\cdot\Phuman(y').\\
    \human&= \argmax_{y \in \mathcal{Y}} u_h(y).  \label{eq:true_MBR decoding}
\end{align}
where utility function $u$: $\mathcal{Y} \times \mathcal{Y} \rightarrow [0,\Umax]$, $\Umax \in [0,1]$ denotes the maximum utility value.

Since $\Phuman$ is unknown, MBR decoding instead uses $\Pmodel\in \mathcal{M}_\mathcal{Y}^{\mathcal{X}}$ to approximate $\Phuman$.
\begin{align}
    u_m(y) &= \sum_{y' \in \mathcal{Y}} u(y, y')\cdot\Pmodel(y').\\
     \hmodel &= \argmax_{y \in \mathcal{Y}} u_m(y).  
\label{eq:model-MBR decoding}
\end{align}
However, summation over $\mathcal{Y}$ is computationally intractable, so Eq.~\eqref{eq:model-MBR decoding} is approximated by a Monte Carlo estimation \cite{eikema-aziz-2022-sampling,farinhas-etal-2023-empirical} using a collection of reference hypotheses $\Yn$ sampled from the model $\Pmodel$.
\begin{align}
    \widehat{u}_m(y) &= \frac{1}{|\Yn|} \sum_{y' \in \Yn} u(y, y').\\
    \hmonte &= \argmax_{y \in \mathcal{\Yn}} \widehat{u}_m(y). \label{eq:monte_MBR decoding}
\end{align}
We denote the number of samples used for the Monte Carlo estimate of the MBR decoding as $n \coloneqq \left|\Yn\right|$.

Therefore, to derive a practical application equation (Eq.~\ref{eq:monte_MBR decoding}), two approximation operations are performed from the objective equation for true MBR decoding (Eq.~\ref{eq:true_MBR decoding}).
\subsection{MAP Decoding}
The most intuitive decoding method is MAP decoding, which selects a mode based on the human distribution $\Phuman$. MAP decoding is also a special case in which the utility function of MBR decoding is used as the indicator function.
The objective function of MAP decoding is defined by:
\begin{equation}\label{eq:map_human}
    \maphuman = \argmax_{y \in \mathcal{Y}} \Phuman(y).
\end{equation}
The objective equation using the model probability $\Pmodel$ is similar to the MBR decoding:
\begin{equation}\label{eq:map_model}
    \mapmodel = \argmax_{y \in \mathcal{Y}} \Pmodel(y).
\end{equation}
In addition, the objective equation for the Monte Carlo estimation of the MAP decoding is defined as:
\begin{equation}\label{eq:empirical_P}
\widehat{P}(y)=\frac{\sum_{{y}^{\prime} \in \Yn} \mathbb{I}\left({y}={y}^{\prime}\right)}{\left|\Yn\right|} .
\end{equation}
where  $\mathcal{Y}^\mathrm{n}_{\mathrm{ref}}$ collected n samples from $\Pmodel$. 

We reformulate the practical objective function of MAP decoding using Eq.~\eqref{eq:empirical_P}:
\begin{equation}\label{eq:map_monte}
    \hm = \argmax_{y \in \Yn} \widehat{P}(y).
\end{equation}
Eq.~\eqref{eq:map_monte} shows the computationally feasible approximation of the MAP decoding. 
While beam search is the most common sampling strategy to approximate MAP decoding. 

We focus on the analysis of MAP decoding and MBR decoding with random sampling in this paper, the case of considering temperature sampling for MBR decoding in Appendix~\ref{appendix:coro_tempre}.
The goal of the study is to investigate the statistics of the MBR and MAP objectives.

\section{Analysis of MBR decoding}
In this section, we evaluate the performance of MBR decoding (Eq.~\ref{eq:monte_MBR decoding}) compared to the ideal MBR solution (Eq.~\ref{eq:true_MBR decoding}) under various assumptions.

\subsection{Problem Setting}
The optimal MBR decoding output is $\human$ (Eq.~\ref{eq:true_MBR decoding}). However, as mentioned earlier, due to practical limitations, only a Monte Carlo solution $\hmonte$ (Eq.~\ref{eq:monte_MBR decoding}) can be obtained in practice. On the other hand, since this $\hmonte$ is ultimately evaluated under the human distribution $\Phuman$, the following performance difference arises:
\begin{equation}
\mathrm{Regret}_{n}\coloneqq u_h(\human) - u_h(\hmonte).
\end{equation}
We refer to this quantity $\mathrm{Regret}_{n}$ as regret.
The goal of this study is to obtain an upper bound of regret theoretically. Suppose we can show this upper bound on the order of the number of elements in candidate and hypothesis sets. In that case, we can provide a theoretical guarantee for the performance of MBR decoding using the Monte Carlo method.

\subsection{The Analysis of $\mathrm{Regret}_{n}$}
We define the following notation:
\begin{equation}\label{eq:delta}
    \Delta(u_p,u_q,y) \coloneqq u_p(y)- u_q(y).
\end{equation}
This expresses the residual of the utility of $y$ assuming $q$ as the probability distribution when the target probability distribution is $p$. 
Using Eq.~\eqref{eq:delta}, we divide the regret $\mathrm{Regret}_{n}$ into four terms:
\begin{align}
 \mathrm{Regret}_{n} &\leq  \Delta(u_h,u_m,\human) + \Delta(u_m,\widehat{u}_m,\human)\nonumber\\
 &+\Delta(\widehat{u}_m,u_m,\hmonte)+ \Delta(u_m,u_h,\hmonte).\label{equation:regret_n}
\end{align}

In the following analysis, we first derive an upper bound for each $\Delta$ in Eq.~\eqref{equation:regret_n}. Then, using the upper bounds derived for each $\Delta$, we derive an upper bound for $\mathrm{Regret}_{n}$.
\paragraph{Analysis of $u_m$ and $\widehat{u}_m$.} 
First, we derive an upper bound for the terms involving $u_m$ (marginalization under $\Pmodel$), $\widehat{u}_m$ (marginalization under Monte Carlo sampling from $\Pmodel$).
We consider the following assumption about the utility function.
\begin{mainassumption}[Inner Product Representation of the Utility Function]\label{assumption:utility}
Let $\ba (y)\in\mathbb{R}^d$ be an embedding for each $y \in \mathcal{Y}$.
We assume that $\max \|\ba\|= \Umax$ and the utility function $u(y, y')$ is given by the inner product of the embeddings, i.e., 
\begin{equation*}
    u(y, y') = \ba (y)^{\top} \ba(y').
\end{equation*}
\end{mainassumption}
There are examples of utility functions that satisfy these properties such as the $F_1$ measure of the BERTScore and the inner product of the embedding functions \cite{bert-score,reimers-gurevych-2019-sentence,SFR-embedding-2}. Note that several state-of-the-art utility functions for machine translation do not satisfy this assumption (e.g., COMET and Metric-X; \citealt{rei-etal-2020-comet, guerreiro-etal-2024-xcomet, juraska-etal-2024-metricx}), since they are trained to do the computation of the utility and the quality estimation at the same time.

By applying Hoeffding’s Inequality (Lemma~\ref{theorem:hoeffding}) and Uniform Concentration Inequality (Lemma~\ref{theorem:uci}), see Appendix~\ref{appendix:lemmas}, along with Assumption~\ref{assumption:utility}, the following lemma about the terms involving $u_m, \widehat{u}_m$ is established.
\begin{mainlemma}[Upper Bound for the terms involving $u_m, \widehat{u}_m$]\label{lemma:heart}
    Under Assumption~\ref{assumption:utility} and assuming $d \geq 4$, the following bound holds for any $\delta\in(0,1)$, with probability at least $1 - \delta$:
\begin{align*}
        &\Delta(u_m,\widehat{u}_m,\human)+\Delta(\widehat{u}_m,u_m,\hmonte) \\
        &\leq 3\sqrt{\frac{1}{n} \log \frac{1}{\delta}} +\frac{36}{n}\sqrt{d \log d}.
    \end{align*}
\end{mainlemma}

The proof can be found in Appendix~\ref{apendix:heart}. Since the dimensions of the embedding models are usually larger than $4$, we assume them in this study and proceed with our analysis under this assumption.\footnote{For readers interested in the case $d < 4$, see Appendix~\ref{apendix:heart}.}
Lemma~\ref{lemma:heart} shows that the upper bound of regret with $u_m, \widehat{u}_m$ terms depends only on the number of samples $n$ and decreases at a rate of $\smash{O}\left(n^{-\frac{1}{2}}\right)$.  
Notably, this result can also be interpreted as a regret bound, specifically $\mathrm{Regret}_{n}$, under the condition that $\Phuman$ and $\Pmodel$ are identical (Appendix~\ref{appendix:human_equal_model}).

Our current analysis relies on the inner product representation assumption for the utility function in order to derive the tight upper bounds presented in our work. In particular, when bounding $\Delta\left(\hat{u}^m, u_m, \hat{y}^m\right)$ in Lemma~\ref{lemma:heart}, we utilize results from \citet{shalev2014understanding}, which allows us to maintain an $\smash{O}\left(n^{-1}\right)$ convergence rate. This result is used in the following analysis (Theorem~\ref{theorem:bound3} and Theorem~\ref{theorem:bound}).


\paragraph{Note.}
Metrics based on neural architectures do not satisfy Assumption~\ref{assumption:utility}. 
Therefore, we introduce a new assumption about the utility function under which these metrics comply.
\begin{mainassumption}\label{assumption:kernel}
    We assume that the utility function is a positive-definite kernel.
\end{mainassumption}
A utility function $u$ is a positive-definite kernel if and only if all kernel matrices resulting from this kernel function are
symmetric positive semidefinite (Definition 7.1 \citet{bach2024learning}).
 Under Assumption~\ref{assumption:kernel}, we can get the following lemma related to the upper bound for the terms involving $u_m, \widehat{u}_m$.

\begin{mainlemma}[Upper Bound for the terms involving $u_m, \widehat{u}_m$]\label{lemma:kernel}
 Under Assumption~\ref{assumption:kernel}, the following bound holds for any $\delta\in(0,1)$, with probability at least $1 - \delta$:
\begin{align*}    &\Delta(u_m,\widehat{u}_m,\human)+\Delta(\widehat{u}_m,u_m,\hmonte)\\
&\leq 3\sqrt{\frac{1}{n} \log \frac{1}{\delta}} +\frac{2}{\sqrt{n}}.
\end{align*}
\end{mainlemma}

The proof of Lemma~\ref{lemma:kernel} is in Appendix~\ref{appendix:kernel}.

Assumption~\ref{assumption:kernel} may not be applicable in the real world. There are cases in which the utility function $u$ does not satisfy semi-positive definiteness and symmetry (e.g., COMET). In these cases, we perform an operation to make the following lemma applicable. First, we convert the asymmetry into symmetry.
\begin{equation*}
    u'(y, y') = \frac{u(y, y') + u(y', y)}{2}.
\end{equation*}
Next, we use Multidimensional Scaling (Section 3.1 \citet{pub:39177}) so that $u'$ satisfies semi-positive definiteness, we can finally apply Lemma~\ref{lemma:kernel}.
From here on, the main section proceeds under Assumption~\ref{assumption:utility}, although the corresponding results under Assumption~\ref{assumption:kernel} follow immediately.
\paragraph{Analysis of $u_h$ and $u_m$.} 
Next, we analyze the $u_h$ (marginalization under $\Phuman$),$u_m$ (marginalization under $\Pmodel$) terms involved, but before doing so, we consider the following assumptions.
\begin{mainassumption}[Utility Function Smoothness]\label{assumption:lip}
    For all $y,y^\prime,y^{\prime \prime} \in \mathcal{Y}$, we assume the utility function satisfies the following inequality:
\begin{equation*}
|u(y,y') - u(y,y'')| \leq C(y',y'')
\end{equation*}
where $C \in \mathbb{R}^{\mathcal{Y}\times \mathcal{Y}}$ is a cost function.
\end{mainassumption}
The assumption is not a restrictive assumption. For any utility functions bounded by $[0, \Umax]$, $C(y', y'') = \Umax$ satisfies the assumption. Note also that Assumption~\ref{assumption:utility} entails Assumption~\ref{assumption:lip}.
The assumption is known as the Lipschitz condition \cite{jeffreys1988lipschitz}. 
It claims that the value of the utility function is smooth under the cost function $C$: the difference in the utility between an output $y$ and other outputs $y'$ and $y''$ can be bounded by some ``distance'' $C$ between the $y'$ and $y''$.
Intuitively, if $y'$ and $y''$ are similar, then $C$ wants to be small, and otherwise large. Many of the utility functions are designed to be so by minimizing the prediction error from the human evaluation (e.g., MQM score) \cite{rei-etal-2020-comet,juraska-etal-2024-metricx}.
Under the Assumption~\ref{assumption:lip}, the following lemma holds: 
\begin{mainlemma}[Upper Bound for the terms involving $u_h, u_m$]\label{lemma:wd} Under Assumption~\ref{assumption:lip}, the following bound can be derived:
\begin{align*}
        &\Delta(u_h,u_m,\human) +\Delta(u_m,u_h,\hmonte) \\
        &\leq 2\mathrm{WD}(P_{\mathrm{human}}, P_{\mathrm{model}}),
    \end{align*}
    where $\mathrm{WD}$ is Wasseratein distance with $C$ being the cost function.
    \end{mainlemma}
The definition of Wasserstein distance \cite{wang2012coupling} is described in Appendix~\ref{appendix:lemmas}. The proof can be found in Appendix~\ref{appendix:wd}.
Lemma~\ref{lemma:wd} implies that minimizing the terms involving  $u_h, u_m$ requires choosing $\Pmodel$ that closely approximates $\Phuman$.

\paragraph{Upper bound of $\mathrm{Regret}_{n}$.}
Using Lemma~\ref{lemma:heart} and Lemma~\ref{lemma:wd}, we can derive an upper bound for $\mathrm{Regret}_{n}$.
\begin{maintheorem}[Regret Bound for MBR decoding]\label{theorem:bound3}
    Under Assumption~\ref{assumption:utility}, Assumption~\ref{assumption:lip}, and assuming $d\geq 4$, the regret upper bound of the MBR decoding holds for any $\delta\in(0,1)$, with probability at least $1 - \delta$:
 \begin{align*}
    &\mathrm{Regret}_{n} \leq  3 \sqrt{\frac{1}{n} \log \frac{1}{\delta}} +\frac{36}{n}\sqrt{d \log d}\\&+2\mathrm{WD}(P_{\mathrm{human}}, P_{\mathrm{model}}).
\end{align*}
\end{maintheorem}
Theorem~\ref{theorem:bound3} can be interpreted as follows. 
First, the upper bound decreases with a larger number of samples from $\Pmodel$ with the convergence speed of $\smash{O}\left(n^{-\frac{1}{2}}\right)$.
This implies that you can reduce the upper bound by $30\%$ by doubling the number of samples $2n$, you are probably to need at least four times more samples $4n$ to reduce the initial error by $50\%$.
The other insight is that the error is inherently limited by the Wasserstein distance between $P_{\mathrm{human}}$ and $P_{\mathrm{model}}$, which means that, as expected, the accuracy of $P_{\mathrm{model}}$ is desirable.
This observation is consistent with the finding that \citet{ohashi-etal-2024-true} indicates that MBR decoding performance is improved when $\Phuman$ and $\Pmodel$ are similar.
\subsection{On the Effect of the Training Dataset Size}
In practice, we cannot compute the exact value of the Wasserstein distance as it requires enumeration over all possible sentences.  
To derive a more digestible bound, we consider the simplest example where $\Pmodel$ is an empirical distribution of $\Phuman$.
Formally, we consider the following assumption:
\begin{mainassumption}[$\Pmodel$ as an Empirical Distribution Sampled the Size of Training Dataset $|D|$ from $\Phuman$.]\label{assumption:human}
    Let $\Pmodel$ be the empirical distribution of $|D|$ samples obtained from $\Phuman$.
$\Pmodel$ has the following expression:
\begin{align*}
    \Pmodel(y) &=  \frac{1}{|D|}\sum_{y^\prime \in D}\mathbb{I}(y= y^\prime). \\
    & D \sim \Phuman(\cdot) 
\end{align*}
where $\mathbb{I}$ is an indicator function.
\end{mainassumption}
Assumption~\ref{assumption:human} is not intended to be an assumption that is directly applicable to real-world scenarios. 
In a real-world scenario, $\Pmodel$ is almost always represented by function approximation models (e.g., neural networks) for text generation tasks. 
They are often pretrained by unsupervised learning using a language model objective, then fine-tuned by supervised learning and preference learning \cite{gpt1,stiennon2020learning,ouyang2022training}.

Given the diversity and complexity of models used in practice, we instead analyze a simple model where it has no function approximation, pretraining, or post-training processes.
Such a simple model is likely to be worse than models used in practice.
Therefore, the bounds we derive from this simple model serve as informal worst-case bounds for the state-of-the-art models.

The size of the training dataset is usually much larger than the number of samples for MBR decoding: $|D| \gg n $. 

\paragraph{Analysis of $u_m$ and $\widehat{u}_m$ with the training dataset size $|D|$.} 
 Under Assumption~\ref{assumption:human}, we derive the analysis on the terms in $u_h,u_m$, using the Hoeffding’s Inequality (Lemma~\ref{theorem:hoeffding}, see Appendix~\ref{appendix:lemmas}), we can get the following the upper bound.
 \begin{mainlemma}[Upper Bound for the terms involving $u_h, u_m$ with the Size of Training Dataset $|D|$]\label{lemma:black}
    Under Assumption ~\ref{assumption:human}, the following bound holds for any $\delta\in(0,1)$, with probability at least $1 - \delta$:
    \begin{align*}
        &\Delta(u_h,u_m,\human)+ \Delta(u_m,u_h,\hmonte)\\
        &\leq 3\sqrt{\frac{1}{|D|} \log \frac{1}{\delta}}.
    \end{align*}
\end{mainlemma}
The proof can be found in Appendix~\ref{appendix:black}.
This shows that the upper bounds for the $u_h,u_m$ terms vary only with the size of the training dataset $|D|$ and that the upper bound decays at a rate of $\smash{O}\left(|D|^{-\frac{1}{2}}\right)$ with its size.

Under Assumption~\ref{assumption:human}, regret depends on both samples $n$ and $|D|$. For clarity, we define a regret under Assumption~\ref{assumption:human} as follows:
\begin{equation}
    \mathrm{Regret}_{n,D}\coloneqq u_h(\human) - u_h(\hmonte).
\end{equation}
\paragraph{Upper bound of $\mathrm{Regret}_{n,D}$.}
We can immediately derive the upper bound for $\mathrm{Regret}_{n,D}$ from Lemma~\ref{lemma:heart} and Lemma~\ref{lemma:black} as follows:
\begin{maintheorem}[Regret Bound for MBR decoding with the Size of Training Dataset $|D|$]\label{theorem:bound}
    Under Assumption~\ref{assumption:utility}, Assumption~\ref{assumption:human}, and assuming $d\geq 4$, the regret upper bound of the MBR decoding holds for any $\delta\in(0,1)$, with probability at least $1 - \delta$:
    \begin{align*}
       \mathrm{Regret}_{n,D}  &\leq 4 \sqrt{\frac{1}{n} \log \frac{1}{\delta}}+4\sqrt{\frac{1}{|D|} \log \frac{1}{\delta}} \\
        &+ \frac{36 }{n}\sqrt{d \log  d}. 
    \end{align*}
\end{maintheorem}
Theorem~\ref{theorem:bound} shows that MBR decoding approaches the optimal output with a convergence rate related to the size of the reference hypothesis set $n$ and the size of the training dataset $|D|$, suggesting why MBR decoding has good experimental performance. 

This implies that even if in the future language models are trained on 1000 times more training data than those today, MBR decoding is likely to be a valid option, given that it scales with the quality of the language model. It also scales with the advancement of hardware accelerators - if we can generate more samples at a time, it will make MBR decoding more effective. MBR decoding is not an algorithm that happens to be useful under current situations of language models and machine translation tasks. It is likely to be a valid option in the future with more sophisticated language models and hardware.

\subsection{Extended Analysis of MBR Decoding}
\paragraph{Expected regret bounds.}
So far, we have found that we can obtain upper bounds that occur with high probability, and from these upper bounds, we can immediately determine the expected regret upper bound for Theorem~\ref{theorem:bound3} and Theorem~\ref{theorem:bound}.
\begin{maincoro}[Expected Regret Upper Bound of MBR decoding]\label{propotion:mbr}
The expected regret upper bounds $\mathrm{Regret}_{n},\mathrm{Regret}_{n,D}$ are bounded as follows for any $\delta\in(0,1)$ under assuming $d\geq4$:
\begin{align*}
    \E\left[\mathrm{Regret}_{n}  \right] &\leq 3 \sqrt{\frac{1}{n} \log \frac{1}{\delta}} +\frac{36}{n}\sqrt{d \log d}\\&+2\mathrm{WD}(P_{\mathrm{human}}, P_{\mathrm{model}}) + \delta 
\end{align*}
\vspace{-0.5cm}
\begin{align*}
    \E\left[\mathrm{Regret}_{n,D}  \right] &\leq 4 \sqrt{\frac{1}{n} \log \frac{1}{\delta}}+\frac{36}{n}\sqrt{d \log  d} \\
        &+4\sqrt{\frac{1}{|D|} \log \frac{1}{\delta}}  + \delta 
\end{align*}
\end{maincoro}
The proof is in Appendix~\ref{appendix:reg_exp}. Corollary~\ref{propotion:mbr} can be used to estimate how large the regret will be, on average.

\paragraph{On the effect of errors in the utility function.}
In the real world, we cannot always have access to the true utility function $u$, instead, we assume that the proxy utility function is only available $u'$, and we explain the bound of the difference in the utility function. We focus exclusively on the conditions outlined in Theorem~\ref{theorem:bound}.

We define the following expression:
\begin{align*}
    u'(y) &= \frac{1}{|\Yn|} \sum_{y' \in \Yn} u'(y, y').\\
    y' &= \argmax_{y \in \mathcal{\Yn}} u'(y). 
\end{align*}
We want to find the upper bound of $u_h(\human) - u_h(y')$.  
Our new objective function in this paragraph is defined as:
\begin{equation}
    \mathrm{Regret}^{u}_{n,D} \coloneqq u_h(\human) - u_h(y').
\end{equation}
Under Assumption~\ref{assumption:utility} and Assumption~\ref{assumption:human}, an upper bound of $\mathrm{Regret}^{u}_{n,D}$ is derived using the Hoeffding's inequality (Lemma~\ref{theorem:hoeffding}). Let $\alpha_{err} \coloneqq \max_{y, y' \in \Yn}|| \ba(y) - \ba'(y')||$.

\begin{maincoro}[Regret Bound for MBR decoding with utility function error]\label{coro:utility}
    Under Assumption~\ref{assumption:utility} and Assumption~\ref{assumption:human}, the regret upper bound of the MBR decoding with utility function error holds for any $\delta\in(0,1)$, with probability at least $1 - \delta$:
\begin{align*}
    \mathrm{Regret}^{u}_{n,D} &\leq 4\sqrt{\frac{1}{|D|} \log \frac{1}{\delta}} + 4 \sqrt{\frac{1}{n} \log \frac{1}{\delta}}\\ &+ 2d  \alpha_{err}.
\end{align*}
\end{maincoro}
The proof can be found in Appendix~\ref{appendix:utility}.
In Corollary~\ref{coro:utility}, the upper bound decreases as the number of samples $n$ and the size of training dataset $|D|$ increases.  
However, it does not ultimately converge to zero, as the term $\alpha_{err}$ remains.

\section{Analysis of MAP Decoding}
In this section,  we derive the regret of MAP decoding between the optimal value and the Monte Carlo estimated value, expressed as $\Phuman(\maphuman) - \Phuman(\hm)$. 

We define the regret of the MAP decoding as follows:
\begin{equation}
\mathrm{Regret}^\mathrm{MAP}_n = \Phuman(\maphuman) - \Phuman(\hm). 
\end{equation}
We refer to $\mathrm{Regret}^\mathrm{MAP}_n$ as MAP regret.
Under the conditions of Theorem~\ref{theorem:bound3}, the upper bound of $\mathrm{Regret}^\mathrm{MAP}_n$ is given by the following result using Dvoretzky–Kiefer–Wolfowitz inequality (Lemma~\ref{theorem:DKW}, see in Appendix~\ref{appendix:lemmas}).
\begin{maintheorem}[Regret Bound for MAP decoding]\label{theorem:map_bound2}
Under the conditions of Theorem~\ref{theorem:bound3}, the MAP regret upper bound of the MAP decoding holds for any $\delta\in(0,1)$, with probability at least $1 - \delta$:
\begin{align*}
    \mathrm{Regret}^{\mathrm{MAP}}_n &\leq 6\sqrt{\frac{1}{n}\log\frac{1}{\delta}} \\
    & + 2\mathrm{WD}(P_{\mathrm{human}}, P_{\mathrm{model}}).
\end{align*}
\end{maintheorem}

Furthermore, under the conditions of Theorem~\ref{theorem:bound}, MAP regret depends on the number of samples $n$ and the size of the training dataset $|D|$.

Our new objective formulation is defined as:
\begin{equation}
\mathrm{Regret}^\mathrm{MAP}_{n,D} = \Phuman(\maphuman) - \Phuman(\hm)
\end{equation}
The upper bound of $\mathrm{Regret}^\mathrm{MAP}_{n,D}$ is also immediately obtained by Lemma~\ref{theorem:DKW} as follows.
\begin{maintheorem}[Regret Bound for MAP decoding with the Size of Training Dataset $|D|$]\label{theorem:map_bound3}
Under the conditions of Theorem~\ref{theorem:bound}, the MAP regret upper bound of the MAP decoding holds for any $\delta\in(0,1)$, with probability at least $1 - \delta$:
\begin{align*}
    \mathrm{Regret}^\mathrm{MAP}_{n,D} &\leq 8\sqrt{\frac{1}{n}\log\frac{1}{\delta}}+8\sqrt{\frac{1}{|D|}\log\frac{1}{\delta}}.
\end{align*}
\end{maintheorem}
The proof is in Appendix~\ref{appendix:map_true}. Note that Theorem~\ref{theorem:map_bound2} and Theorem~\ref{theorem:map_bound3} decrease in the same order as Theorem~\ref{theorem:bound3} and Theorem~\ref{theorem:bound} respectively.
In other words, if we compare the difference between these bounds more clearly, we focus on the constant term.

\section{Performance Comparison}\label{section:map_mbr}
So far, we have analyzed MBR decoding and MAP decoding independently.  
In this section, we compare the performance of MBR decoding and MAP decoding within the same framework, in terms of the upper bound and we focus exclusively on the conditions outlined in Theorem~\ref{theorem:bound}, Theorem~\ref{theorem:map_bound3}.

\paragraph{Difference between MBR and MAP Decoding target values.}
First, we aim to analyze the difference $u_h(\human) - u_h(\hm) $, where $\human$ is the optimal output and $\hm$ is a suboptimal output. We can analyze this error bound to see how the optimal solution in MAP decoding behaves in an ideal MBR decoding environment.
\begin{mainobservation}[Error between $\human$ and $\hm$ with $u_h$]\label{obs:1}
    Error bound between $\human$ and $\hm$ with $u_h$ satisfies for any $\delta\in\left(0,\frac{2}{5}\right)$, with probability at least $1-\frac{5}{2}\delta$:
\begin{align*}
     &u_h(\human) -u_h(\hm) \leq  u_m(\hmonte) -u_m(\hm) \\
     &+ \smash{O}\left(\max \left(n^{-\frac{1}{2}}, |D|^{-\frac{1}{2}}\right)\right).
\end{align*}
\end{mainobservation}
The detail is in Appendix~\ref{appendix:observation1}.
This observation confirms that MAP decoding and MBR decoding theoretically have different objectives under certain conditions. 
We provide the analysis of the MAP regret $\mathrm{Regret}^\mathrm{MAP}_{n,D}$ of the two decoding algorithms in Appendix~\ref{appendix:observation1}.

\paragraph{Convergence speed of upper bound of $\mathrm{Regret}_{n,D}$ and $\mathrm{Regret}^\mathrm{MAP}_{n,D}$.}
Next, we compare the upper bounds of the convergence rate between MBR decoding and MAP decoding presented in this study.
\begin{mainobservation}[Comparison of the Convergence Speed]\label{obs:2}
We compare the upper bounds of $\mathrm{Regret}_{n,D}$ and $\mathrm{Regret}^\mathrm{MAP}_{n,D}$ under three different scenarios:
\paragraph{1. $n \to \infty$ and $|D|$ is finite.}
The upper bound of $\mathrm{Regret}_{n,D}$ is strictly smaller than the upper bound of $\mathrm{Regret}^\mathrm{MAP}_{n,D}$.

\paragraph{2. $D \to \infty$ and $n$ is finite.}
For number of samples $n$ and utility $d$ such that the following inequality holds, the upper bound of $\mathrm{Regret}_{n,D}$ is smaller than the upper bound of $\mathrm{Regret}^\mathrm{MAP}_{n,D}$.
 \begin{align*}
        \frac{1}{9}\sqrt{n\log{\frac{1}{\delta}}} \geq \sqrt{d \log d}.
    \end{align*}
    
\paragraph{3. Both $D$ and $n$ are finite.}
For the number of samples $n$, utility $d$, and the size of training dataset $|D|$ such that the following inequality holds, the upper bound of $\mathrm{Regret}_{n,D}$ is smaller than the upper bound of $\mathrm{Regret}^\mathrm{MAP}_{n,D}$.
\begin{align*}
    \frac{n}{9}\left(\sqrt{\frac{1}{n} \log \frac{1}{\delta}}+\sqrt{\frac{1}{|D|} \log \frac{1}{\delta}}\right) &\geq \sqrt{d \log d}.
\end{align*}
\end{mainobservation}
\begin{figure}[t]
    \centering
    \includegraphics[width=\linewidth]{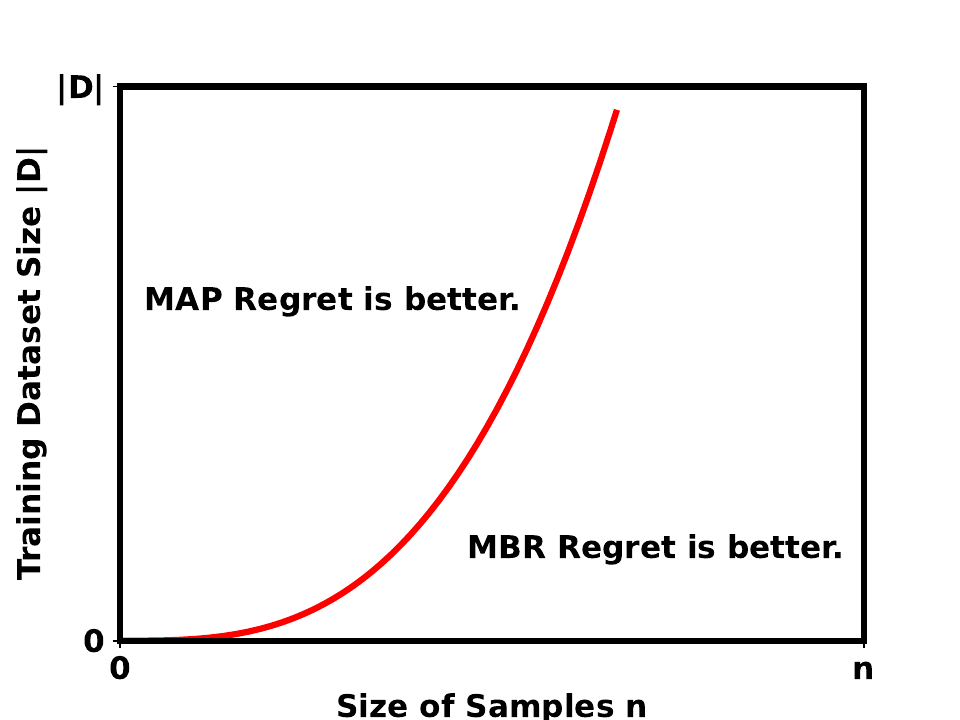}
    \caption{Conceptual visualization of Observation~\ref{obs:2}. The convergence rates of the upper bound of $\mathrm{Regret}_{n, D}$ and $\mathrm{Regret}_{n, D}^{\mathrm{MAP}}$ with the number of samples $n$ and training dataset size $|D|$ are compared. For $n$ and $|D|$ on the right side of this line plot, it means that the upper bound of $\mathrm{Regret}_{n, D}$ is smaller.}
    \label{fig:compare}
\end{figure}
The proofs are given in Appendix~\ref{appendix:observation2}, and Fig.~\ref{fig:convergence-rate} shows that the upper bound of MBR decoding is less than the upper bound of MAP decoding at the number of samples described in Observation~\ref{obs:2}'s 3.

\begin{figure}
    \centering
    \includegraphics[width=\linewidth]{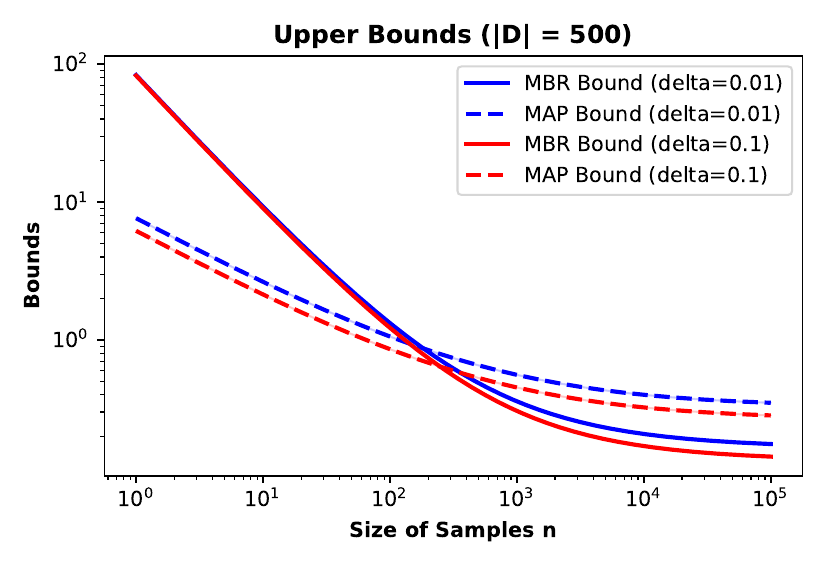}
    \caption{Numerical Experiment for Observation~\ref{obs:2}'s 3}
    \label{fig:convergence-rate}
\end{figure}

As can be seen by comparing $\mathrm{Regret}_{n, D}$ and $\mathrm{Regret}_{n, D}^{\mathrm{MAP}}$ in cases 2 and 3 of Observation~\ref{obs:2}, as the number of samples $n$ increases, the upper bound on $\mathrm{Regret}_{n, D}$ converges more faster (Fig.~\ref{fig:compare}). 

The analysis shows that, for any model, there exists a large enough $n$ such that the upper bound of the regret of MBR decoding is smaller than that of MAP decoding. 
This observation may help explain why the empirical performance of MBR decoding can exceed that of MAP decoding.


\section{Numerical Simulation}\label{sec:experiments}
In this section, we computationally evaluate the upper bounds of $\mathrm{Regret}_{n}$ and $\mathrm{Regret}_{n,D}$.  
It is important to emphasize that the experiments conducted in this paper are not intended to show the tightness of the results in the practical NLP tasks. Rather, they are intended to provide a visual representation of the theoretical results presented in this paper.

For the performance of MBR decoding in real-world NLP tasks, we refer to previous work \cite{freitag-etal-2023-epsilon,bertsch-etal-2023-mbr,heineman-etal-2024-improving,wu2025better}.

We consider $\mathcal{Y}=10{,}000$ hypotheses $y_i$ ($i=1,\dots,\mathcal{H}$) with the dataset size of $|D|$ and the number of samples $n$ model samples to study regret and bound behavior. We test $\delta\in\{0.01,0.1\}$, and set $d=4$. 
The details of the experimental setup are given in Appendix~\ref{appendix:exp}.
\subsection{Results}
 \begin{figure}
             \centering
             \includegraphics[width=\linewidth]{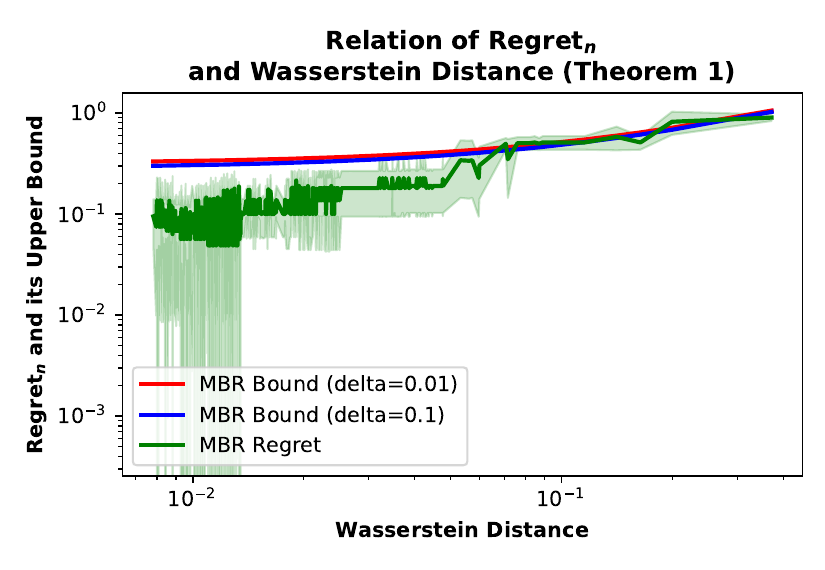}
             \caption{The upper bound ($\delta=\{\color{red}0.01,\color{blue}0.1\color{black}\}$) of the $\mathrm{Regret}_{n}$ derived by Theorem~\ref{theorem:bound3} and the value of \color{teal} $\mathrm{Regret}_{n}$\color{black} in the simulation. The number of samples $n$ is fixed to $500$ and the training dataset size is $|D|=[0,1000]$}
             \label{fig:wasserstein_distance}
         \end{figure}
\begin{figure}[t]
         \includegraphics[width=0.48\textwidth]{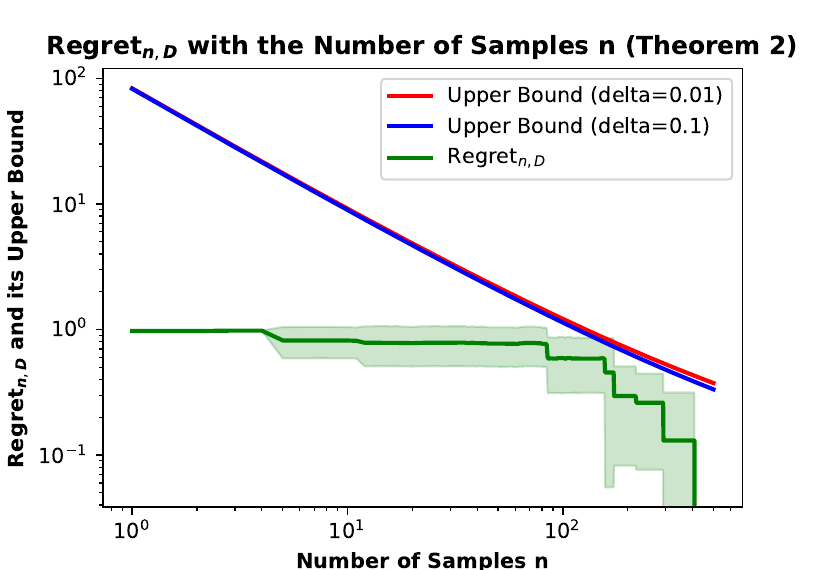}
         \caption{The upper bound ($\delta=\{\color{red}0.01,\color{blue}0.1\color{black}\}$) of the $\mathrm{Regret}_{n,D}$ derived by Theorem~\ref{theorem:bound} and the value of \color{teal} $\mathrm{Regret}_{n,D}$\color{black} in the simulation. The training dataset size $|D|$ is fixed to $5000$ and the number of samples for MBR decoding is $n=[0,500]$.}
         \label{fig:fix_d}
         \end{figure}
         
         \begin{figure}
             \includegraphics[width=\linewidth]{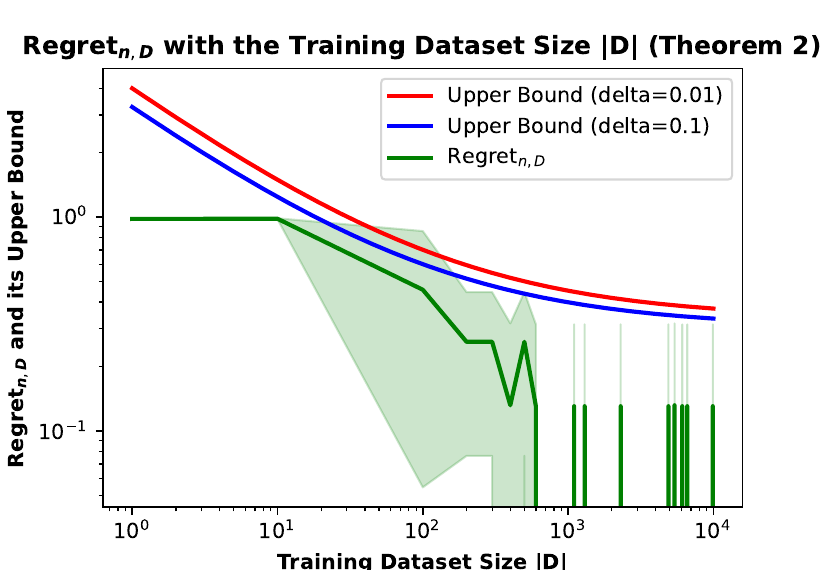}
             \caption{The upper bound ($\delta=\{\color{red}0.01,\color{blue}0.1\color{black}\}$) of the $\mathrm{Regret}_{n,D}$ derived by Theorem~\ref{theorem:bound} and the value of \color{teal} $\mathrm{Regret}_{n,D}$ \color{black} in simulation. The number of samples $n$ is fixed to $500$ and the training dataset size is $|D|=[0,10000]$}
             \label{fig:fix_n}
         \end{figure}

Fig.~\ref{fig:wasserstein_distance} clearly demonstrates that our theoretical upper bound on $\mathrm{Regret}_{n}$ is tight when compared to the actual regret observed. This close correspondence indicates that the assumptions and inequalities used in deriving the bound are well-justified, providing evidence in the numerical experiment. 

The results of Fig.~\ref{fig:fix_d} and Fig.~\ref{fig:fix_n} show that the obtained upper bound converges to $\mathrm{Regret}_{n,D}$ as the number of samples increases. 
This behavior suggests the theoretical validity of the bound indicating that the looseness of the upper bound is gradually eliminated as the number of samples increases, improving the ability to accurately capture the true performance difference accurately. 

\section{Conclusions}
This paper presents a theoretical analysis of MBR decoding and shows that, under reasonable assumptions, it converges with high probability to the optimal solution at a rate of $\smash{O}\left(n^{-\frac{1}{2}}\right)$, even when the total language space $\mathcal{Y}$ is large. 
In addition, we compare MBR and MAP decoding about the performance difference and the convergence speed to the optimal solution. 
As a result, we confirm MAP decoding and MBR decoding theoretically have different objectives, and from the upper bound, MBR decoding is more efficient than MAP decoding in approaching the optimal output under certain conditions. 


\section{Limitations}
\label{sec:limitations}

This study provides the first theoretical bounds on MBR decoding. As it is one of the first analyses on MBR decoding, it has several limitations, particularly regarding its alignment with practical implementations.


\paragraph{Assumptions.}
Our analysis assumes that the set of candidate hypotheses $\mathcal{H}$ is identical to the set of reference outputs $\mathcal{Y}$. However, in practice, using a small number of high-quality but biased candidates alongside a larger set of unbiased references has been found to be more effective.

We have considered three assumptions for the analysis. The assumptions do not cover all the situations of text generation applications. For example, the state-of-the-art utility functions for machine translation (COMET and Metric-X; \citealt{rei-etal-2020-comet,guerreiro-etal-2024-xcomet,juraska-etal-2024-metricx}) are not linear function (Assumption~\ref{assumption:utility}).

In practice, the models are represented by a neural network, and they are often pretrained using unsupervised learning before supervised fine-tuning. This point is not considered in Assumption~\ref{assumption:human}.

\paragraph{Aspects not considered.}
The analysis does not account for the role of neural networks. In particular, it is known that solutions corresponding to flat minima tend to generalize better than those with sharp minima \cite{dinh2017sharp}. Understanding the role of neural networks for MBR decoding is future work.

We analyze a model that predicts sequences, but practical implementations typically use autoregressive language models \cite{lin-etal-2021-limitations}. Incorporating the autoregressive assumption may lead to improved theoretical bounds.

The study considers only random sampling and temperature sampling. However, other strategies, such as epsilon sampling \cite{hewitt-etal-2022-truncation} and beam search (which is commonly used for MAP decoding), are not analyzed.

Our analysis does not frame the problem as an NLP task. Incorporating domain-specific characteristics could lead to tighter bounds.
This study is purely theoretical and does not include empirical experiments to validate the results on real-world NLP tasks. Instead, we rely on prior experimental findings \cite{freitag-etal-2023-epsilon,bertsch-etal-2023-mbr,suzgun-etal-2023-follow} for providing empirical support for our theoretical conclusions.

Another key limitation is that the bounds derived in this study are not proven to be tight, leaving room for refinement. Furthermore, to measure how tight the upper bounds is, we also need to derive the lower bound in MBR decoding.

Lastly, while our study focuses on sample complexity, practical implementations of MBR decoding are often constrained by computational complexity \cite{cheng-vlachos-2023-faster,vamvas2024lineartime}. Combining our sampling complexity result with the existing computational complexity bounds \cite{jinnai2024hyperparameterfree} is future work.

\paragraph{Summary.}
This work provides fundamental theoretical bounds for MBR decoding. However, there remain avenues for improvement, including empirical validation, refinement of theoretical bounds, and comparative analysis with alternative decoding algorithms.


\section{Ethics Statements}
\label{sec:ethics}
We do not foresee any ethical concerns regarding the analysis of the paper.
\section*{Acknowledgments}
Kaito Ariu is supported by JSPS KAKENHI Grant No. 23K19986.

\bibliography{anthology,ms}

\begin{thebibliography}{47}
\expandafter\ifx\csname natexlab\endcsname\relax\def\natexlab#1{#1}\fi

\bibitem[{Bach(2024)}]{bach2024learning}
Francis Bach. 2024.
\newblock \href {https://books.google.co.jp/books?id=R_T8EAAAQBAJ} {\emph{Learning {T}heory from {F}irst {P}rinciples}}.
\newblock Adaptive Computation and Machine Learning series. MIT Press.

\bibitem[{Bertsch et~al.(2023)Bertsch, Xie, Neubig, and Gormley}]{bertsch-etal-2023-mbr}
Amanda Bertsch, Alex Xie, Graham Neubig, and Matthew Gormley. 2023.
\newblock \href {https://doi.org/10.18653/v1/2023.bigpicture-1.9} {It{'}s {MBR} all the way down: Modern generation techniques through the lens of minimum {B}ayes risk}.
\newblock In \emph{Proceedings of the Big Picture Workshop}, pages 108--122, Singapore. Association for Computational Linguistics.

\bibitem[{Borgeaud and Emerson(2020)}]{borgeaud-emerson-2020-leveraging}
Sebastian Borgeaud and Guy Emerson. 2020.
\newblock \href {https://doi.org/10.18653/v1/2020.ngt-1.11} {Leveraging sentence similarity in natural language generation: Improving beam search using range voting}.
\newblock In \emph{Proceedings of the Fourth Workshop on Neural Generation and Translation}, pages 97--109, Online. Association for Computational Linguistics.

\bibitem[{Cheng and Vlachos(2023)}]{cheng-vlachos-2023-faster}
Julius Cheng and Andreas Vlachos. 2023.
\newblock \href {https://aclanthology.org/2023.emnlp-main.767} {Faster minimum {B}ayes risk decoding with confidence-based pruning}.
\newblock In \emph{Proceedings of the 2023 Conference on Empirical Methods in Natural Language Processing}, pages 12473--12480, Singapore. Association for Computational Linguistics.

\bibitem[{de~Gispert et~al.(2009)de~Gispert, Virpioja, Kurimo, and Byrne}]{de-gispert-etal-2009-minimum}
Adri{\`a} de~Gispert, Sami Virpioja, Mikko Kurimo, and William Byrne. 2009.
\newblock \href {https://aclanthology.org/N09-2019} {Minimum {B}ayes risk combination of translation hypotheses from alternative morphological decompositions}.
\newblock In \emph{Proceedings of Human Language Technologies: The 2009 Annual Conference of the North {A}merican Chapter of the Association for Computational Linguistics, Companion Volume: Short Papers}, pages 73--76, Boulder, Colorado. Association for Computational Linguistics.

\bibitem[{Dinh et~al.(2017)Dinh, Pascanu, Bengio, and Bengio}]{dinh2017sharp}
Laurent Dinh, Razvan Pascanu, Samy Bengio, and Yoshua Bengio. 2017.
\newblock Sharp minima can generalize for deep nets.
\newblock In \emph{International Conference on Machine Learning}, pages 1019--1028. PMLR.

\bibitem[{Ehling et~al.(2007)Ehling, Zens, and Ney}]{ehling-etal-2007-minimum}
Nicola Ehling, Richard Zens, and Hermann Ney. 2007.
\newblock \href {https://aclanthology.org/P07-2026} {Minimum {B}ayes risk decoding for {BLEU}}.
\newblock In \emph{Proceedings of the 45th Annual Meeting of the Association for Computational Linguistics Companion Volume Proceedings of the Demo and Poster Sessions}, pages 101--104, Prague, Czech Republic. Association for Computational Linguistics.

\bibitem[{Eikema and Aziz(2020)}]{eikema-aziz-2020-map}
Bryan Eikema and Wilker Aziz. 2020.
\newblock \href {https://doi.org/10.18653/v1/2020.coling-main.398} {Is {MAP} decoding all you need? the inadequacy of the mode in neural machine translation}.
\newblock In \emph{Proceedings of the 28th International Conference on Computational Linguistics}, pages 4506--4520, Barcelona, Spain (Online). International Committee on Computational Linguistics.

\bibitem[{Eikema and Aziz(2022)}]{eikema-aziz-2022-sampling}
Bryan Eikema and Wilker Aziz. 2022.
\newblock \href {https://doi.org/10.18653/v1/2022.emnlp-main.754} {Sampling-{B}ased {A}pproximations to {M}inimum {B}ayes {R}isk {D}ecoding for {N}eural {M}achine {T}ranslation}.
\newblock In \emph{Proceedings of the 2022 Conference on Empirical Methods in Natural Language Processing}, pages 10978--10993, Abu Dhabi, United Arab Emirates. Association for Computational Linguistics.

\bibitem[{Farinhas et~al.(2023)Farinhas, de~Souza, and Martins}]{farinhas-etal-2023-empirical}
Ant{\'o}nio Farinhas, Jos{\'e} de~Souza, and Andre Martins. 2023.
\newblock \href {https://doi.org/10.18653/v1/2023.emnlp-main.733} {An {E}mpirical {S}tudy of {T}ranslation {H}ypothesis {E}nsembling with {L}arge {L}anguage {M}odels}.
\newblock In \emph{Proceedings of the 2023 Conference on Empirical Methods in Natural Language Processing}, pages 11956--11970, Singapore. Association for Computational Linguistics.

\bibitem[{Fernandes et~al.(2022)Fernandes, Farinhas, Rei, De~Souza, Ogayo, Neubig, and Martins}]{fernandes-etal-2022-quality}
Patrick Fernandes, Ant{\'o}nio Farinhas, Ricardo Rei, Jos{\'e} De~Souza, Perez Ogayo, Graham Neubig, and Andre Martins. 2022.
\newblock \href {https://doi.org/10.18653/v1/2022.naacl-main.100} {Quality-aware decoding for neural machine translation}.
\newblock In \emph{Proceedings of the 2022 Conference of the North American Chapter of the Association for Computational Linguistics: Human Language Technologies}, pages 1396--1412, Seattle, United States. Association for Computational Linguistics.

\bibitem[{Freitag et~al.(2023)Freitag, Ghorbani, and Fernandes}]{freitag-etal-2023-epsilon}
Markus Freitag, Behrooz Ghorbani, and Patrick Fernandes. 2023.
\newblock \href {https://doi.org/10.18653/v1/2023.findings-emnlp.617} {Epsilon sampling rocks: Investigating sampling strategies for minimum {B}ayes risk decoding for machine translation}.
\newblock In \emph{Findings of the Association for Computational Linguistics: EMNLP 2023}, pages 9198--9209, Singapore. Association for Computational Linguistics.

\bibitem[{Freitag et~al.(2022)Freitag, Grangier, Tan, and Liang}]{freitag-etal-2022-high}
Markus Freitag, David Grangier, Qijun Tan, and Bowen Liang. 2022.
\newblock \href {https://doi.org/10.1162/tacl_a_00491} {High quality rather than high model probability: Minimum {B}ayes risk decoding with neural metrics}.
\newblock \emph{Transactions of the Association for Computational Linguistics}, 10:811--825.

\bibitem[{Groenen and Borg(2013)}]{pub:39177}
Patrick Groenen and Ingwer Borg. 2013.
\newblock The {P}ast, {P}resent, and {F}uture of {M}ultidimensional {S}caling.
\newblock \emph{Report / Econometric Institute, Erasmus University Rotterdam}, pages 1--25.

\bibitem[{Guerreiro et~al.(2024)Guerreiro, Rei, Stigt, Coheur, Colombo, and Martins}]{guerreiro-etal-2024-xcomet}
Nuno~M. Guerreiro, Ricardo Rei, Daan~van Stigt, Luisa Coheur, Pierre Colombo, and Andr{\'e} F.~T. Martins. 2024.
\newblock \href {https://doi.org/10.1162/tacl_a_00683} {xcomet: Transparent machine translation evaluation through fine-grained error detection}.
\newblock \emph{Transactions of the Association for Computational Linguistics}, 12:979--995.

\bibitem[{Heineman et~al.(2024)Heineman, Dou, and Xu}]{heineman-etal-2024-improving}
David Heineman, Yao Dou, and Wei Xu. 2024.
\newblock \href {https://doi.org/10.18653/v1/2024.emnlp-main.1255} {Improving {M}inimum {B}ayes {R}isk {D}ecoding with {M}ulti-{P}rompt}.
\newblock In \emph{Proceedings of the 2024 Conference on Empirical Methods in Natural Language Processing}, pages 22525--22545, Miami, Florida, USA. Association for Computational Linguistics.

\bibitem[{Hewitt et~al.(2022)Hewitt, Manning, and Liang}]{hewitt-etal-2022-truncation}
John Hewitt, Christopher Manning, and Percy Liang. 2022.
\newblock \href {https://doi.org/10.18653/v1/2022.findings-emnlp.249} {Truncation sampling as language model desmoothing}.
\newblock In \emph{Findings of the Association for Computational Linguistics: EMNLP 2022}, pages 3414--3427, Abu Dhabi, United Arab Emirates. Association for Computational Linguistics.

\bibitem[{Jeffreys and Jeffreys(1988)}]{jeffreys1988lipschitz}
H~Jeffreys and BS~Jeffreys. 1988.
\newblock The {L}ipschitz {C}ondition.
\newblock \emph{Methods of Mathematical Physics}, page~53.

\bibitem[{Jinnai and Ariu(2024)}]{jinnai2024hyperparameterfree}
Yuu Jinnai and Kaito Ariu. 2024.
\newblock \href {https://doi.org/10.18653/v1/2024.findings-acl.505} {Hyperparameter-free approach for faster minimum {B}ayes risk decoding}.
\newblock In \emph{Findings of the Association for Computational Linguistics: ACL 2024}, pages 8547--8566, Bangkok, Thailand. Association for Computational Linguistics.

\bibitem[{Juraska et~al.(2024)Juraska, Deutsch, Finkelstein, and Freitag}]{juraska-etal-2024-metricx}
Juraj Juraska, Daniel Deutsch, Mara Finkelstein, and Markus Freitag. 2024.
\newblock \href {https://doi.org/10.18653/v1/2024.wmt-1.35} {{M}etric{X}-24: The {G}oogle {S}ubmission to the {WMT} 2024 {M}etrics {S}hared {T}ask}.
\newblock In \emph{Proceedings of the Ninth Conference on Machine Translation}, pages 492--504, Miami, Florida, USA. Association for Computational Linguistics.

\bibitem[{Kamigaito et~al.(2024)Kamigaito, Deguchi, Sakai, Hayashi, and Watanabe}]{kamigaito2024theoretical}
Hidetaka Kamigaito, Hiroyuki Deguchi, Yusuke Sakai, Katsuhiko Hayashi, and Taro Watanabe. 2024.
\newblock Theoretical {A}spects of {B}ias and {D}iversity in {M}inimum {B}ayes {R}isk {D}ecoding.
\newblock \emph{arXiv preprint arXiv:2410.15021}.

\bibitem[{Kumar and Byrne(2002)}]{kumar-byrne-2002-minimum}
Shankar Kumar and William Byrne. 2002.
\newblock \href {https://doi.org/10.3115/1118693.1118712} {Minimum {B}ayes-risk word alignments of bilingual texts}.
\newblock In \emph{Proceedings of the 2002 Conference on Empirical Methods in Natural Language Processing ({EMNLP} 2002)}, pages 140--147. Association for Computational Linguistics.

\bibitem[{Kumar and Byrne(2004)}]{kumar-byrne-2004-minimum}
Shankar Kumar and William Byrne. 2004.
\newblock \href {https://aclanthology.org/N04-1022} {Minimum {B}ayes-risk decoding for statistical machine translation}.
\newblock In \emph{Proceedings of the Human Language Technology Conference of the North {A}merican Chapter of the Association for Computational Linguistics: {HLT}-{NAACL} 2004}, pages 169--176, Boston, Massachusetts, USA. Association for Computational Linguistics.

\bibitem[{Lin et~al.(2021)Lin, Jaech, Li, Gormley, and Eisner}]{lin-etal-2021-limitations}
Chu-Cheng Lin, Aaron Jaech, Xin Li, Matthew~R. Gormley, and Jason Eisner. 2021.
\newblock \href {https://doi.org/10.18653/v1/2021.naacl-main.405} {Limitations of autoregressive models and their alternatives}.
\newblock In \emph{Proceedings of the 2021 Conference of the North American Chapter of the Association for Computational Linguistics: Human Language Technologies}, pages 5147--5173, Online. Association for Computational Linguistics.

\bibitem[{Massart(1990)}]{massart1990tight}
Pascal Massart. 1990.
\newblock The tight constant in the {D}voretzky-{K}iefer-{W}olfowitz inequality.
\newblock \emph{The annals of Probability}, pages 1269--1283.

\bibitem[{Meng et~al.(2024)Meng, Liu, Joty, Xiong, Zhou, and Yavuz}]{SFR-embedding-2}
Rui Meng, Ye~Liu, Shafiq~Rayhan Joty, Caiming Xiong, Yingbo Zhou, and Semih Yavuz. 2024.
\newblock \href {https://huggingface.co/Salesforce/SFR-Embedding-2_R} {S{FR}-{E}mbedding-2: {A}dvanced {T}ext {E}mbedding with {M}ulti-stage {T}raining}.

\bibitem[{Mohri(2018)}]{mohri2018foundations}
Mehryar Mohri. 2018.
\newblock Foundations of machine learning.

\bibitem[{M{\"u}ller and Sennrich(2021)}]{muller-sennrich-2021-understanding}
Mathias M{\"u}ller and Rico Sennrich. 2021.
\newblock \href {https://doi.org/10.18653/v1/2021.acl-long.22} {Understanding the properties of minimum {B}ayes risk decoding in neural machine translation}.
\newblock In \emph{Proceedings of the 59th Annual Meeting of the Association for Computational Linguistics and the 11th International Joint Conference on Natural Language Processing (Volume 1: Long Papers)}, pages 259--272, Online. Association for Computational Linguistics.

\bibitem[{Narayan et~al.(2018)Narayan, Cohen, and Lapata}]{narayan-etal-2018-dont}
Shashi Narayan, Shay~B. Cohen, and Mirella Lapata. 2018.
\newblock \href {https://doi.org/10.18653/v1/D18-1206} {Don{'}t give me the details, just the summary! topic-aware convolutional neural networks for extreme summarization}.
\newblock In \emph{Proceedings of the 2018 Conference on Empirical Methods in Natural Language Processing}, pages 1797--1807, Brussels, Belgium. Association for Computational Linguistics.

\bibitem[{Ohashi et~al.(2024)Ohashi, Honda, Morimura, and Jinnai}]{ohashi-etal-2024-true}
Atsumoto Ohashi, Ukyo Honda, Tetsuro Morimura, and Yuu Jinnai. 2024.
\newblock \href {https://doi.org/10.18653/v1/2024.naacl-short.38} {On the {T}rue {D}istribution {A}pproximation of {M}inimum {B}ayes-{R}isk {D}ecoding}.
\newblock In \emph{Proceedings of the 2024 Conference of the North American Chapter of the Association for Computational Linguistics: Human Language Technologies (Volume 2: Short Papers)}, pages 459--468, Mexico City, Mexico. Association for Computational Linguistics.

\bibitem[{Ouyang et~al.(2022)Ouyang, Wu, Jiang, Almeida, Wainwright, Mishkin, Zhang, Agarwal, Slama, Ray et~al.}]{ouyang2022training}
Long Ouyang, Jeffrey Wu, Xu~Jiang, Diogo Almeida, Carroll Wainwright, Pamela Mishkin, Chong Zhang, Sandhini Agarwal, Katarina Slama, Alex Ray, et~al. 2022.
\newblock Training language models to follow instructions with human feedback.
\newblock \emph{Advances in neural information processing systems}, 35:27730--27744.

\bibitem[{Peyré and Cuturi(2020)}]{peyre2020computational}
Gabriel Peyré and Marco Cuturi. 2020.
\newblock Computational optimal transport.
\newblock \emph{arXiv preprint arXiv:1803.00567}.

\bibitem[{Radford et~al.(2018)Radford, Narasimhan, Salimans, and Sutskever}]{gpt1}
Alec Radford, Karthik Narasimhan, Tim Salimans, and Ilya Sutskever. 2018.
\newblock Improving language understanding by generative pre-training.

\bibitem[{Rei et~al.(2020)Rei, Stewart, Farinha, and Lavie}]{rei-etal-2020-comet}
Ricardo Rei, Craig Stewart, Ana~C Farinha, and Alon Lavie. 2020.
\newblock \href {https://doi.org/10.18653/v1/2020.emnlp-main.213} {{COMET}: A neural framework for {MT} evaluation}.
\newblock In \emph{Proceedings of the 2020 Conference on Empirical Methods in Natural Language Processing (EMNLP)}, pages 2685--2702, Online. Association for Computational Linguistics.

\bibitem[{Reimers and Gurevych(2019)}]{reimers-gurevych-2019-sentence}
Nils Reimers and Iryna Gurevych. 2019.
\newblock \href {https://doi.org/10.18653/v1/D19-1410} {Sentence-{BERT}: Sentence embeddings using {S}iamese {BERT}-networks}.
\newblock In \emph{Proceedings of the 2019 Conference on Empirical Methods in Natural Language Processing and the 9th International Joint Conference on Natural Language Processing (EMNLP-IJCNLP)}, pages 3982--3992, Hong Kong, China. Association for Computational Linguistics.

\bibitem[{Rush et~al.(2015)Rush, Chopra, and Weston}]{rush-etal-2015-neural}
Alexander~M. Rush, Sumit Chopra, and Jason Weston. 2015.
\newblock \href {https://doi.org/10.18653/v1/D15-1044} {A neural attention model for abstractive sentence summarization}.
\newblock In \emph{Proceedings of the 2015 Conference on Empirical Methods in Natural Language Processing}, pages 379--389, Lisbon, Portugal. Association for Computational Linguistics.

\bibitem[{Shalev-Shwartz and Ben-David(2014)}]{shalev2014understanding}
Shai Shalev-Shwartz and Shai Ben-David. 2014.
\newblock \emph{Understanding machine learning: {F}rom theory to algorithms}.
\newblock Cambridge university press.

\bibitem[{Stahlberg et~al.(2017)Stahlberg, de~Gispert, Hasler, and Byrne}]{stahlberg-etal-2017-neural}
Felix Stahlberg, Adri{\`a} de~Gispert, Eva Hasler, and Bill Byrne. 2017.
\newblock \href {https://aclanthology.org/E17-2058} {Neural machine translation by minimising the {B}ayes-risk with respect to syntactic translation lattices}.
\newblock In \emph{Proceedings of the 15th Conference of the {E}uropean Chapter of the Association for Computational Linguistics: Volume 2, Short Papers}, pages 362--368, Valencia, Spain. Association for Computational Linguistics.

\bibitem[{Stiennon et~al.(2020)Stiennon, Ouyang, Wu, Ziegler, Lowe, Voss, Radford, Amodei, and Christiano}]{stiennon2020learning}
Nisan Stiennon, Long Ouyang, Jeffrey Wu, Daniel Ziegler, Ryan Lowe, Chelsea Voss, Alec Radford, Dario Amodei, and Paul~F Christiano. 2020.
\newblock Learning to summarize with human feedback.
\newblock \emph{Advances in Neural Information Processing Systems}, 33:3008--3021.

\bibitem[{Suzgun et~al.(2023)Suzgun, Melas-Kyriazi, and Jurafsky}]{suzgun-etal-2023-follow}
Mirac Suzgun, Luke Melas-Kyriazi, and Dan Jurafsky. 2023.
\newblock \href {https://doi.org/10.18653/v1/2023.findings-acl.262} {Follow the wisdom of the crowd: Effective text generation via minimum {B}ayes risk decoding}.
\newblock In \emph{Findings of the Association for Computational Linguistics: ACL 2023}, pages 4265--4293, Toronto, Canada. Association for Computational Linguistics.

\bibitem[{Tromble et~al.(2008)Tromble, Kumar, Och, and Macherey}]{tromble-etal-2008-lattice}
Roy Tromble, Shankar Kumar, Franz Och, and Wolfgang Macherey. 2008.
\newblock \href {https://aclanthology.org/D08-1065} {Lattice {M}inimum {B}ayes-{R}isk decoding for statistical machine translation}.
\newblock In \emph{Proceedings of the 2008 Conference on Empirical Methods in Natural Language Processing}, pages 620--629, Honolulu, Hawaii. Association for Computational Linguistics.

\bibitem[{Vamvas and Sennrich(2024)}]{vamvas2024lineartime}
Jannis Vamvas and Rico Sennrich. 2024.
\newblock Linear-time minimum bayes risk decoding with reference aggregation.
\newblock \emph{arXiv preprint arXiv:2402.04251}.

\bibitem[{Wainwright(2019)}]{wainwright2019high}
Martin~J Wainwright. 2019.
\newblock \emph{High-dimensional statistics: {A} non-asymptotic viewpoint}, volume~48.
\newblock Cambridge university press.

\bibitem[{Wang(2012)}]{wang2012coupling}
Feng-Yu Wang. 2012.
\newblock Coupling and applications.
\newblock In \emph{Stochastic Analysis and Applications to Finance: Essays in Honour of Jia-An Yan}, pages 411--424. World Scientific.

\bibitem[{Wu et~al.(2025)Wu, Fernandes, Bertsch, Kim, Pakazad, and Neubig}]{wu2025better}
Ian Wu, Patrick Fernandes, Amanda Bertsch, Seungone Kim, Sina~Khoshfetrat Pakazad, and Graham Neubig. 2025.
\newblock \href {https://openreview.net/forum?id=7xCSK9BLPy} {Better {I}nstruction-{F}ollowing {T}hrough {M}inimum {B}ayes {R}isk}.
\newblock In \emph{The Thirteenth International Conference on Learning Representations}.

\bibitem[{Yan et~al.(2024)Yan, Xu, Meng, Zhou, and Zhang}]{yan-etal-2024-dc}
Jianhao Yan, Jin Xu, Fandong Meng, Jie Zhou, and Yue Zhang. 2024.
\newblock \href {https://aclanthology.org/2024.lrec-main.395/} {{DC}-{MBR}: {D}istributional {C}ooling for {M}inimum {B}ayesian {R}isk {D}ecoding}.
\newblock In \emph{Proceedings of the 2024 Joint International Conference on Computational Linguistics, Language Resources and Evaluation (LREC-COLING 2024)}, pages 4423--4437, Torino, Italia. ELRA and ICCL.

\bibitem[{Zhang et~al.(2020)Zhang, Kishore, Wu, Weinberger, and Artzi}]{bert-score}
Tianyi Zhang, Varsha Kishore, Felix Wu, Kilian~Q. Weinberger, and Yoav Artzi. 2020.
\newblock \href {https://openreview.net/forum?id=SkeHuCVFDr} {Bertscore: Evaluating text generation with bert}.
\newblock In \emph{International Conference on Learning Representations}.

\end{thebibliography}
\newpage
\appendix
\onecolumn

\section{Related Works}


\paragraph{Experimental findings in MBR decoding.}
Many studies have reported that the performance of MBR decoding increases with a larger number of samples \cite{eikema-aziz-2022-sampling,freitag-etal-2022-high}.
Prior works \cite{freitag-etal-2022-high,fernandes-etal-2022-quality} show that the performance of the MBR decoding depends on the selection of the utility function. 
Experiments combining MBR decoding with neural reference-based metrics, such as BLEURT, demonstrate significant improvements in human evaluations. 
In recent work, \citet{yan-etal-2024-dc} propose Distributional Cooling MBR, this approach bridges the gap between label smoothing and MBR decoding, with extensive experimental validation demonstrating its effectiveness on various NMT benchmarks and \citet{wu2025better} shows that leveraging reference-based LLM judges with MBR decoding improves the output quality of instruction-following LLMs compared to greedy decoding, best-of-N approaches.

\paragraph{Analysis of MBR decoding.}
\citet{kamigaito2024theoretical} conduct on the intricate relationship between bias and diversity in MBR decoding. Their bias-diversity decomposition framework theoretically explains the trade-offs observed in empirical studies.
\section{Useful Lemmas and Definition}
\label{appendix:lemmas}
In this section, we present the concentration inequality used in the paper.
The following inequalities represent a uniform concentration inequality.

\begin{lemma}[Hoeffding's inequality; Corollary 1.1 in \citealt{bach2024learning}]\label{theorem:hoeffding}
$\{X_i\}_{i=1}^n \in [0,b]$ being i.i.d. samples drawn from same distribution.
    \begin{equation*}
    \operatorname{Pr}\left(\left|\E\left[X\right] - \frac{1}{n}\sum_{i=1}^n X_i\right| \geq \epsilon\right) 
    \leq 2\exp\left(-\frac{2 n\epsilon^2}{b^2}\right).
\end{equation*}
\end{lemma}
The following inequalities represent a uniform concentration inequality.
\begin{lemma}[Uniform Concentration Inequality; Theorem 4.10 in \citealt{wainwright2019high}]\label{theorem:uci}
$\mathcal{F}$ is a class of functions $f: \mathcal{X} \rightarrow [0,b]$.
\begin{align*}
    \operatorname{Pr}\left(\left\|\mathbb{P}_n - \mathbb{P}\right\|_{\mathcal{F}} \geq 2 \mathcal{R}_n(\mathcal{F}) + \epsilon\right) 
    &\leq  \exp\left(-\frac{2n \epsilon^2}{b^2}\right).
\end{align*}
\end{lemma}

where $\left\|\mathbb{P}_n - \mathbb{P}\right\|_{\mathcal{F}} = \sup_{f \in \mathcal{F}} \left|\mathbb{P}_n f - \mathbb{P} f\right|$, $\mathbb{P}_{n} f = \frac{1}{n} \sum_{i=1}^n f(X_i)$ and $\mathbb{P} f = \E[f(X)]$, with $X$ and $\{X_i\}_{i=1}^n$ being i.i.d. samples drawn from $\mathbb{P}$, $\mathcal{R}_n:\left(\mathcal{F},\left\{X_i\right\}_{i=1}^n\right) \rightarrow \mathbb{R}$.

$\mathcal{R}_n(\mathcal{F})$ represents the Rademacher complexity of the function class $\mathcal{F}$ (Definition 3.1 \citep{mohri2018foundations}). Rademacher complexity is a measure of model complexity, indicating how well a function class can fit random noise. It provides a uniform bound on the deviation between the empirical and true expectations across all functions in $\mathcal{F}$, serving as a key tool for analyzing generalization error in statistical learning theory.

\begin{lemma}[Dvoretzky–Kiefer–Wolfowitz inequality; \citealt{massart1990tight}]\label{theorem:DKW}
\begin{equation*}
\operatorname{Pr}\left(\sup _{x \in \mathbb{R}}\left|F_n(x)-F(x)\right|>\varepsilon\right) \leq 2 \exp(-2 n \varepsilon^2 ).
\end{equation*}
\end{lemma}

Given a natural number $n$, let $X_1, X_2, \cdots, X_n$ be real-valued independent and identically distributed random variables with cumulative distribution function $F(\cdot)$. Let $F_n$ denote the associated empirical distribution function defined by $
F_n(x)=\frac{1}{n} \sum_{i=1}^n \mathbf{1}_{\left\{X_i \leq x\right\}}$

\begin{definition}[Wassertstein Distance]
    
The Wassertstein Distance ($\mathrm{WD}$) \citep{wang2012coupling} is defined as:
\begin{equation}\label{eq:wd}
\mathrm{WD}(\nu , \mu) = \inf_{\gamma \in \Gamma(\nu, \mu)} \sum_{(i, j) \in N \times N} \gamma_{ij} \, C_{ij},
\end{equation}
\end{definition}
where $N$: the total number of samples, consisting of the set $\{ y_1, y_2, \dots, y_N \}$, $\nu, \mu \in \Delta_N$: probability measure on the aforementioned sets ($\nu_i, \mu_i$ refer to the probability value $\nu(y_i), \mu(y_i)$), $C$: $N\times N \rightarrow \mathbb{R}$ a cost function measuring the distance between two outputs (e.g. $C_{ij}$ refers to the amount to
be transported from place $y_i$ to palace $y_j$), and $\Gamma(\nu, \mu)$ denotes the set of all joint distributions $\gamma$ whose marginals are $\nu$ and $\mu$. The constraints on $\gamma$ are given by:
\begin{equation*}
\begin{aligned}
\sum_{j \in n} \gamma_{ij} &= \nu_i, \quad \forall i \in n, \\
\sum_{i \in n} \gamma_{ij} &= \mu_j, \quad \forall j \in n, \\
\gamma_{ij} &\geq 0, \quad \forall i,j \in n.
\end{aligned}
\end{equation*}
The $\mathrm{WD}$, also known as the Earth Mover's Distance (EMD), is a metric used to quantify the dissimilarity between two probability distributions.
Unfortunately, computing WD between two probability distributions over $\mathcal{Y}$ exactly is generally intractable, as it requires an enumeration over $\mathcal{Y}$.
Still, WD can be approximated by using empirical distributions with a finite number of samples with the convergence rate of $\smash{O}(n^{-\frac{1}{d}})$ \cite{peyre2020computational}.
\section{Proof of Lemma~\ref{lemma:heart}}
\label{apendix:heart}
We start by analyzing the $u_m(\human) - u_m(\hmonte)$. 

We decompose it as follows:
      \begin{align*}
        &u_m(\human)-\widehat{u}_m(\human) + \widehat{u}_m(\hmonte) - u_m(\hmonte) +\underbrace{\widehat{u}_m(\human) - \widehat{u}_m(\hmonte)}_{\leq 0}\\
        &\leq \Delta(u_m,\widehat{u}_m,\human)+\Delta(\widehat{u}_m,u_m,\hmonte)
    \end{align*}

We can express $ \Delta(u_m,\widehat{u}_m,\human)$ using the following formulation, derived from Lemma~\ref{theorem:hoeffding}.

\begin{equation*}
    \operatorname{Pr}\left(\left| \Delta(u_m,\widehat{u}_m,\human)\right| \leq \epsilon\right) 
    \leq 1- 2\exp\left(-\frac{2 n\epsilon^2}{\Umax^2}\right) = {1- \frac{\delta}{2}}.
\end{equation*}
$ \Delta(u_m,\widehat{u}_m,\human)$ holds the following inequality with probability at least $1-\frac{\delta}{2}$.
\begin{align*}
    \Delta(u_m,\widehat{u}_m,\human)  \leq \Umax\sqrt{\frac{1}{2n}\log \left(\frac{4}{\delta}\right)} .
\end{align*}

Next, we analyze $\Delta(\widehat{u}_m,u_m,\hmonte)$, however, Lemma~\ref{theorem:hoeffding} cannot be directly applied because $\hmonte$ depends on $\widehat{u}_m$. To address this dependency, we instead utilize Lemma~\ref{theorem:uci}, and we can get the following formulation.
\begin{align*}
    \operatorname{Pr}\left(\max_y\left|\Delta(\widehat{u}_m,u_m,y)\right| \leq 2 \mathcal{R}_n(\mathcal{F}) + \epsilon\right) 
    &\leq 1- \exp\left(-\frac{2n \epsilon^2}{\Umax^2}\right) = 1-\frac{\delta}{2}.
\end{align*}
We can thus express $\Delta(\widehat{u}_m,u_m,\hmonte)$ with probability at least $1- \frac{\delta}{2}$.
\begin{align*}
    \Delta(\widehat{u}_m,u_m,\hmonte)\leq \max_y\left|\Delta(\widehat{u}_m,u_m,y)\right| \leq2 \mathcal{R}_n(\mathcal{F}) +\Umax\sqrt{\frac{1}{2n}\log \left(\frac{2}{\delta}\right)}.
\end{align*}

From Section 27.2 \cite{shalev2014understanding}, the following upper bound on the Rademacher complexity $\mathcal{R}_n(\mathcal{F})$ is obtained under Assumption~\ref{assumption:utility}:
\begin{align*}
    2 \mathcal{R}_n(\mathcal{F}) \leq \frac{12 \Umax}{n}\left(\sqrt{d \log (2 \sqrt{d})}+2 \sqrt{d}\right)
\end{align*}
From above inequality, we can get the following the bound:
\begin{align*}
    \Delta(\widehat{u}_m,u_m,\hmonte)&\leq \max_y\left|\Delta(\widehat{u}_m,u_m,y)\right| \\
    &\leq\frac{12 \Umax}{n}\left(\sqrt{d \log (2 \sqrt{d})}+2 \sqrt{d}\right) +\Umax\sqrt{\frac{1}{2n}\log \left(\frac{2}{\delta}\right)}\\
    &\leq\frac{12 \Umax}{n}\left(\sqrt{d \log (2 \sqrt{d})}+2 \sqrt{d}\right) +\Umax\sqrt{\frac{1}{2n}\log \left(\frac{4}{\delta}\right)}.\\
\end{align*}
If $d \geq 4$, 
\begin{align*}
    \Delta(u_m,\widehat{u}_m,\human)+\Delta(\widehat{u}_m,u_m,\hmonte)& \leq\frac{36 \Umax}{n}\sqrt{d \log(\sqrt{d})} +2\Umax\sqrt{\frac{1}{2n}\log \left(\frac{4}{\delta}\right)}.\\
    &\leq 3\sqrt{\frac{1}{n} \log \frac{1}{\delta}} +\frac{36}{n}\sqrt{d \log d}.
\end{align*}
Otherwise, if  $d < 4$,
\begin{align*}
    \Delta(u_m,\widehat{u}_m,\human)+\Delta(\widehat{u}_m,u_m,\hmonte)& \leq\frac{36 \Umax}{n}\sqrt{d \log(\sqrt{d})} +2\Umax\sqrt{\frac{1}{2n}\log \left(\frac{4}{\delta}\right)}.\\
    &\leq 3\sqrt{\frac{1}{n} \log \frac{1}{\delta}} +\frac{72\sqrt{d}}{n}.
\end{align*}
\section{The Case of $\Phuman$ and $\Pmodel$ are identical.}\label{appendix:human_equal_model}
If $\Phuman$ and $\Pmodel$ are equal, the upper bound of $\mathrm{Regret}_{n}$ corresponds to Lemma~\ref{lemma:heart}:
    \begin{align*}
    \mathrm{Regret}_{n} &\leq  2\Umax \sqrt{\frac{1}{2n} \log \frac{8}{\delta}} + \frac{12 \Umax}{n}\left(\sqrt{d \log (2 \sqrt{d})}+2 \sqrt{d}\right).\\
    &\leq 3\sqrt{\frac{1}{n} \log \frac{1}{\delta}} +\frac{36}{n}\left(\sqrt{d \log d}\right).
\end{align*}
In most studies, the primary goal of MBR decoding studies is to derive $\hmodel$, given that $\Phuman$ is inaccessible. These studies implicitly assume $\Pmodel=\Phuman$, highlighting the significance of the results. This finding is for understanding and improving the practical application of MBR decoding methods.

\section{Proof of Lemma~\ref{lemma:wd}}\label{appendix:wd}
We can  derive the following inequality under Assumption~\ref{assumption:lip} for any $y \in \mathcal{Y}$:
\begin{align*}
   \Delta(u_h,u_m,y) &\leq  | \Delta(u_h,u_m,y)|\\
    &= \left|\sum_{y'\in \mathcal{Y}} P_{\mathrm{human}}(y') u(y,y') - \sum_{y'' \in \mathcal{Y}} P_{\mathrm{model}}(y'') u(y,y^{\prime\prime})\right|\\
    &= \left| \sum_{y',y^{\prime\prime}} (u(y,y') - u(y,y^{\prime\prime})) \gamma(y',y^{\prime\prime}) \right| \\
    & \text{where} \quad \sum_{y'} \gamma(y',y^{\prime\prime})= P_{\mathrm{model}}(y^{\prime\prime}), \quad  \sum_{y^{\prime\prime}} \gamma(y,y^{\prime\prime}) = P_{\mathrm{human}}(y'), \quad \gamma(y',y^{\prime\prime}) \geq 0 \\
    &\leq \min_\gamma \sum_{y',y^{\prime\prime}} |u(y,y') - u(y,y^{\prime\prime})| \gamma(y',y^{\prime\prime})  \\
    &\leq \min_\gamma \sum_{y',y^{\prime\prime}} C(y',y^{\prime\prime}) \gamma(y',y^{\prime\prime})  \\
    &= \mathrm{WD}(\Phuman,\Pmodel)
\end{align*}





\section{Proof of Lemma~\ref{lemma:black}}
\label{appendix:black}

Under the Assumption ~\ref{assumption:human}, with using the Lemma ~\ref{theorem:hoeffding}, the $\Delta(u_h,u_m,\human)$ term is expressed as follows:
\begin{equation*}
    \operatorname{Pr}\left(\left|\Delta(u_h,u_m,\human) \right| \leq \epsilon\right) 
    \leq 1- 2\exp\left(-\frac{2 |D|\epsilon^2}{\Umax^2}\right) = {1- \frac{\delta}{2}}.
\end{equation*}
We rearrange $\epsilon$ as follows:

\begin{align*}
    \epsilon &= \Umax\sqrt{\frac{1}{2|D|}\log \left(\frac{4}{\delta}\right)}.
\end{align*}  

In other words, the upper bound of $\Delta(u_h,u_m,\human)$ has a probability of at least $1-\frac{\delta}{2}$.
\begin{align*}
    \Delta(u_h,u_m,\human) \leq \Umax\sqrt{\frac{1}{2|D|}\log \left(\frac{4}{\delta}\right)}.
\end{align*}
For the $\Delta(u_h,u_m,\hmonte)$ term, the upper bound can be obtained by the same operation, and the complement Lemma~\ref{lemma:black} is proved. 

\begin{align*}
        &\Delta(u_h,u_m,\human)+ \Delta(u_m,u_h,\hmonte)\leq 3\sqrt{\frac{1}{|D|} \log \frac{1}{\delta}}.
    \end{align*}

\section{Regret Bound for MBR decoding with temperature sampling}\label{appendix:coro_tempre}

We have been considering random sampling so far, but we also analyze what the bounds would be if we did temperature sampling, considering practical aspects.
\begin{equation*}
    \Pmodel^t(y) = \frac{\exp\left(t^{-1}\Pmodel(y)\right)}{\sum_{y^\prime \in \mathcal{Y}} \exp\left(t^{-1}\Pmodel(y^\prime)\right)}
\end{equation*}
where $t\in\mathbb{R}^{+}$. 
The objective equation for MBR decoding of the Monte Carlo estimates using a collection of reference hypotheses $\Yn$ sampled from the model $\Pmodel^t$ is as follows:
\begin{align*}
    \widehat{u}_m^t(y) &= \frac{1}{|\Yn|} \sum_{y' \in \Yn} u(y, y').\\
\hat{y}^m_t &= \argmax_{y \in \mathcal{\Yn}} \widehat{u}_m^t(y).
\end{align*}
Our new objective formulation is defined as:
\begin{equation}
\mathrm{Regret}_{n,D}^t \coloneqq u_h(\human) - u_h(\hat{y}^m_t).
 \end{equation}
We can derive the upper bound of $\mathrm{Regret}_{n,D}^t$ as follows under the condition Theorem~\ref{theorem:bound} for any $t\in \mathbb{R}^+$.
\begin{maincoro}[Regret Bound for MBR decoding with temperature sampling]\label{coro:bound}
    Under Assumption~\ref{assumption:utility}, Assumption~\ref{assumption:lip}, Assumption~\ref{assumption:human} and assuming $d \geq 4$, the regret upper bound of the MBR decoding holds for any $\delta\in(0,1)$, with probability at least $1 - \delta$:
    \begin{align*}
       &\mathrm{Regret}_{n,D}^t \leq 4 \sqrt{\frac{1}{n} \log \frac{1}{\delta}}+ 4 \sqrt{\frac{1}{|D|} \log \frac{1}{\delta}} \\
        &+ \frac{36}{n}\sqrt{d \log d} + \mathrm{WD}(\Pmodel,\Pmodel^t). 
    \end{align*}
\end{maincoro}
The case of the little samples $n$ might lead to better performance with using the temperature sampling rather than using $\Pmodel$ if $t$ is large. However, the above bound has an extra term $\mathrm{WD}$ when performing temperature sampling compared to Theorem~\ref{theorem:bound}, so the upper bound of $\mathrm{Regret}_{n,D}^t $ might be improved tighter than Corollary~\ref{coro:bound}'s derived from. We discuss this later in this section.
\begin{proof}
    
From the definition of $\mathrm{Regret}_{n,D}^t$:
\begin{equation*}
    \mathrm{Regret}_{n,D}^t \coloneqq u_h(\human) - u_h(\hat{y}^m_t).
 \end{equation*}
 The objective equation for MBR decoding using temperature model distribution can be redefined as:
\begin{align*}
    u^t_m(y) &= \sum_{y' \in \mathcal{Y}} u(y, y')\cdot\Pmodel^t(y').\\
     \hmodel_t &= \argmax_{y \in \mathcal{H}} u_m(y). 
\end{align*}
 We can decompose the following terms:
\begin{align*}
     \mathrm{Regret}_{n,D}^t &\leq  \Delta(u_h,u_m,\human)+\Delta(u_m,u_h,\hmonte_t) + u_m(\human) - u^t_m(\human)+u_m^t(\human) - \widehat{u}^t_m(\human)\\
     &+ \widehat{u}^t_m(\hmonte_t) - u_m^t(\hmonte_t) + u_m^t(\hmonte_t) - u_m(\hmonte_t)\\
     &= \Delta(u_h,u_m,\human)+\Delta(u_m,u_h,\hmonte_t) + \Delta(u_m,u_m^t,\human) + \Delta(u_m^t,\widehat{u}^t_m,\human) \\
     &+ \Delta(\widehat{u}^t_m,u_m^t,\hmonte_t) + \Delta(u_m^t,u_m,\hmonte_t)
\end{align*}

First, the involving the terms $u_h,u_m$ is immediately bounded by Lemma~\ref{lemma:black} with probability at least $1-\frac{\delta}{2}$ under Assumption~\ref{assumption:human}.
\begin{align*}
    \Delta(u_h,u_m,\human) + \Delta(u_m,u_h,\hmonte_t) &\leq 2\Umax\sqrt{\frac{1}{2|D|}\log \left(\frac{8}{\delta}\right)}\\
    &\leq4 \sqrt{\frac{1}{|D|} \log \frac{1}{\delta}}.
\end{align*}

Next we derive $\Delta(u_m,u_m^t,\human) + \Delta(u_m^t,\widehat{u}^t_m,\human)$'s upper bound.
Under Assummption~\ref{assumption:lip}, we can get the following the bound.
\begin{align*}
    \Delta(u_m,u_m^t,\human) +\Delta(u_m^t,u_m,\hmonte_t) \leq 2\mathrm{WD}(\Pmodel,\Pmodel^t)
\end{align*}

Finally, we derive the upper bound for the terms involving $u_m$ and $\widehat{u}^t_m$.  
We conduct the sample operation he proof of Lemma~\ref{lemma:heart}.
We can get the following bound with probability at least $1-\frac{\delta}{2}$ under Assumption~\ref{assumption:utility} and assuming $d\geq 4$:
\begin{align*}
\Delta(u_m^t,\widehat{u}^t_m,\human)+\Delta(\widehat{u}^t_m,u_m^t,\hmonte_t)  
&\leq 2\Umax\sqrt{\frac{1}{2n}\log \left(\frac{8}{\delta}\right)} + \frac{12 \Umax}{n}\left(\sqrt{d \log (2 \sqrt{d})}+2 \sqrt{d}\right)\\
&\leq 4 \sqrt{\frac{1}{n} \log \frac{1}{\delta}}
        + \frac{36 }{n}\sqrt{d \log  d}. \\
\end{align*}
\end{proof}

\paragraph{The upper bound of $\mathrm{Regret}_{n,D}^t $ might be improved}
We focus on $\Delta(u_m,u_m^t,\human)$. In this paper, we can derive the upper bound with Wasserstein Distance.
However, if $\Pmodel$ capture $\Phuman$, $u_m(\human)< u_h(\human)$, but we consider $\Pmodel^t$, it is possible to be $y_m^t = \human$ with little samples, so $\Delta(u_m,u_m^t,\human)$ can be negative value.
In summary, rather than simply deriving an upper bound on the Wasserstein Distance, this bound could be improved by taking into account a more detailed analysis of the temperature sampling.
\section{Proof of the Corollary~\ref{propotion:mbr}}\label{appendix:reg_exp}
We drive the expected upper bound from high probability upper bound. If we have the regret value $R$ with probability at least $1-\delta$, we can get the expected upper bound the following the equation with worst-case value $U$ (e.g. when considering the MBR decoding in this paper, worst-case value can be $1$.)
\begin{align*}
    \textbf{Expected Upper Bound for Regret} = (1-\delta)\cdot R + \delta \cdot U  
\end{align*}
By applying the above equation to Theorem~\ref{theorem:bound3} and Theorem~\ref{theorem:bound}, the following upper bound is derived.
\begin{align*}
    \E\left[\mathrm{Regret}_{n}  \right] &\leq 3 \sqrt{\frac{1}{n} \log \frac{1}{\delta}} +\frac{36}{n}\sqrt{d \log d}+2\mathrm{WD}(P_{\mathrm{human}}, P_{\mathrm{model}}) + \delta 
\end{align*}
\begin{align*}
    \E\left[\mathrm{Regret}_{n,D}  \right] &\leq 4 \sqrt{\frac{1}{n} \log \frac{1}{\delta}}+\frac{36}{n}\sqrt{d \log  d}
        +4\sqrt{\frac{1}{|D|} \log \frac{1}{\delta}}  + \delta 
\end{align*}

\section{Proof of Corollary~\ref{coro:utility}}\label{appendix:utility}
Before the proof, we derive the upper bound of the utility function difference.

The expectation difference is:
\begin{align*}
    &\mathbb{E}\left[u(y, y')\right] 
    \;-\; \mathbb{E}\left[u'(y, y')\right]\\
    &= \mathbb{E}\left[\ba(y)^{\top}\bv(y')\right]
       - \mathbb{E}\left[\ba'(y)^{\top}\bv(y')\right]\\
    &= \mathbb{E}\left[(\,\ba(y)
      - \ba'(y)\,)^{\top} \bv(y')\right].
\end{align*}
By applying the Cauchy–Schwarz inequality, we obtain:
\begin{align*}
    &\E[u(\human,y')] - \E[u^\prime(\human,y')] \\
    &\leq \|\ba(\human)-\ba^{\prime}(\human)\|\cdot \|\E[\bv(y')]\|\\
    &\leq \|\ba(\human)-\ba^{\prime}(\human)\|.
\end{align*}

Next, we prove the corollary~\ref{coro:utility}.
\begin{align*}
    u_h(\human) - u_h(y') = \Delta(u_h,u_m,\human)+ u_m(\human) - u_m(y') + \Delta(u_m,u_h,y').
\end{align*}
From Appendix~\ref{appendix:black}, we can get the following bound with probability at least $1-\frac{\delta}{2}$:
\begin{align*}
    \Delta(u_h,u_m,\human) + \Delta(u_m,u_h,y')\leq 4\sqrt{\frac{1}{|D|} \log \frac{1}{\delta}}.
\end{align*}

The next step is to prove the remaining conditions.
\begin{align*}
    u_m(\human) - u_m(y')&=\Delta(u_m,u',\human) +\Delta(u',u_m,y') - u'(y') + u'(\human)\\
    &\leq \Delta(u_m,u',\human) +\Delta(u',u_m,y')\\
    &= \Delta(u_m,\widehat{u}_m,\human) +\Delta(\widehat{u}_m,u',\human) +\Delta(\widehat{u}_m,u_m,y')+\Delta(u',\widehat{u}_m,y')\\
    &\leq  4 \sqrt{\frac{1}{n} \log \frac{1}{\delta}} + \|\ba(\human)-\ba^{\prime}(\human)\| + \|\ba(y')-\ba^{\prime}(y')\|.
\end{align*}

Finally, we can get the following the bound under Assumption~\ref{assumption:utility} and Assumption~\ref{assumption:human} with probability at least $1-\delta$:
\begin{align*}
    \mathrm{Regret}^{u}_{n,D} \leq 4\sqrt{\frac{1}{|D|} \log \frac{1}{\delta}} + 4 \sqrt{\frac{1}{n} \log \frac{1}{\delta}} + \|\ba(\human)-\ba^{\prime}(\human)\| + \|\ba(y')-\ba^{\prime}(y')\|.
\end{align*}

\section{MAP Decoding Upper Bound} \label{appendix:map_true}

In MAP decoding, our objective is to analyze the difference $\Phuman(\maphuman) - \Phuman(\hm)$.  
To obtain an upper bound, we decompose $\Phuman(\maphuman) - \Phuman(\hm)$ as follows.
\begin{align*}
\Phuman(\maphuman) - \Phuman(\hm) &= \Phuman(\maphuman) - \Pmodel(\maphuman)
+\Pmodel(\maphuman) - \Pmodel(\hm)\\
&+ \Pmodel(\hm) - \Phuman(\hm).
\end{align*}

We solve the upper bound with Lemma~\ref{theorem:DKW}, so we bound the difference of distributions with the difference of empirical distributions.
We also denote $y^{-}$ as the value immediately before $y$. 

The following equation holds for all $y$ with probability at least $1-\frac{2}{\delta}$:
\begin{align*}
|\widehat{P}(y)-\Pmodel(y)|&=\left|\left(\widehat{F}(y)-\widehat{F}\left(y^{-}\right)\right)-\left(F_{\mathrm{model}}(y)-F_{\mathrm{model}}\left(y^{-}\right)\right)\right| .\\
&\leq \left(|\widehat{F}(y)-F_{\mathrm{model}}(y)|+\left|\widehat{F}\left(y^{-}\right)-F_{\mathrm{model}}\left(y^{-}\right)\right|\right) .\\
\max _y|\widehat{P}(y)-\Pmodel(y)|&\leq \max _y\left(|\widehat{F}(y)-F_{\mathrm{model}}(y)|+\left|\widehat{F}\left(y^{-}\right)-F_{\mathrm{model}}\left(y^{-}\right)\right|\right)\label{eq:F_P} \leq 2\epsilon_1 .\\
\end{align*}


We apply Lemma~\ref{theorem:DKW} to the above formulation:
    \begin{equation*}
\operatorname{Pr}\left( \max_{y \in \mathcal{Y}} \left|\widehat{F}(y)-F_{\mathrm{model}}(y)\right|>\epsilon_1\right) \leq 2 \exp \left(-2 n\epsilon_1^2\right).
\end{equation*}

Finally, we get the bound with probability at least $1-\frac{\delta}{2}$:
\begin{align*}
    \max _{y \in \mathcal{Y}}|\widehat{P}(y)-\Pmodel(y)|&\leq 2 \sqrt{\frac{1}{2n}\log \frac{8}{\delta}}.
\end{align*}

The following inequality holds for $\hm$:
\begin{equation*}
    \Pmodel(\hm) \geq \widehat{P}(\hm)-2\epsilon_1.
\end{equation*}

This also applies to $\maphuman$:
\begin{equation*}
    \Pmodel\left(\maphuman\right) \leq \widehat{P}\left(\maphuman\right)+2\epsilon_1.
\end{equation*}

From the definition, it is clear that $\widehat{P}(\hm) \geq\widehat{P}\left(\maphuman\right)$.
\begin{align*}
    \Pmodel(\hm) &\geq \widehat{P}(\hm)-2\epsilon_1 \\
    &\geq \widehat{P}\left(\maphuman\right)-2\epsilon_1 \\
    &\geq \Pmodel\left(\maphuman\right)-4 \epsilon_1 .
\end{align*}

Therefore, we can get the upper bound at least $1-\frac{\delta}{2}$:
\begin{align*}
    \Pmodel\left(\maphuman\right)-\Pmodel(\hm) &\leq 4\sqrt{\frac{1}{2n}\log \frac{8}{\delta}}.
\end{align*}
\begin{equation*}
\operatorname{Pr}\left( \max_{y \in \mathcal{Y}} \left|F_{\mathrm{model}}(h)-F_{\mathrm{human}}(h)\right|>\epsilon_2\right) \leq 2 \exp \left(-2 |D|\epsilon_2^2\right).
\end{equation*}

We also use Lemma~\ref{theorem:DKW}. It satisfies with probability at least $1-\frac{\delta}{2}$:
\begin{align*}
        \Phuman(\maphuman) - \Pmodel(\maphuman) +  \Pmodel(\hm) - \Phuman(\hm)&\leq 4\sqrt{\frac{1}{2|D|}\log \frac{8}{\delta}}.
    \end{align*}

Finally,  we get the following upper bound with probability at least $1-\delta$:

\begin{align*}
   \mathrm{Regret}^\mathrm{MAP}_{n,D}&\leq 4\sqrt{\frac{1}{2n}\log \frac{8}{\delta}} + 4\sqrt{\frac{1}{2|D|}\log \frac{8}{\delta}}\\
    &\leq 8\sqrt{\frac{1}{n}\log\frac{1}{\delta}}+8\sqrt{\frac{1}{|D|}\log\frac{1}{\delta}}.
\end{align*}

\section{Observation}\label{appendix:map_mbr}
We describe the derivations of the Observation~\ref{obs:1} and \ref{obs:2}.

\subsection{Proof of Observation 1}\label{appendix:observation1}

Assuming the MBR decoding goal is the true value, we aim to know $\Phuman(h^*) - \Phuman(\hmonte)$, where $\hmonte$ is the optimal probability based on the empirical distribution of $\Pmodel$, $u_h(h) = \sum \Phuman(y)u(h,y)$.
Remind of $u_h(\human) - u_h(\hmonte)\leq 4 \sqrt{\frac{1}{n} \log \frac{1}{\delta}}+4\sqrt{\frac{1}{|D|} \log \frac{1}{\delta}} 
        + \frac{36 }{n}\sqrt{d \log  d} = \sigma_1$
\begin{align*}
    \Phuman(\hm)u_h(\human) - \Phuman(\hm)u_h(\hmonte) &\leq \Phuman(\hm) \cdot \sigma_1.
\end{align*}

Remind of $\Phuman(h^*) - \Phuman(\hm)\leq 4\left(\sqrt{\frac{1}{n}} + \sqrt{\frac{1}{|D|}}\right)\left(\sqrt{\frac{1}{2}\log\frac{8}{\delta}} \right)$.
\begin{align*}
    \Phuman(h^*)u_h(\hm) - \Phuman(\hm)u_h(\hm)&\leq \underbrace{8\sqrt{\frac{1}{n}\log\frac{1}{\delta}}+8\sqrt{\frac{1}{|D|}\log\frac{1}{\delta}}}_{\sigma_2}.
\end{align*}

Combined above formulation:

\begin{align*}
    \Phuman(\hm)u_h(\human) &- \Phuman(\hm)u_h(\hmonte) + \Phuman(h^*)u_h(\hm) - \Phuman(\hm)u_h(\hm)\\&\leq  \Phuman(\hm) \sigma_1 +\sigma_2,\\
    &\Phuman(\hm)u_h(\human)- \Phuman(\hm)u_h(\hm) \\
    &\leq \Phuman(\hm)u_h(\hmonte)-\Phuman(h^*)u_h(\hm)+ \Phuman(\hm) \sigma_1 +\sigma_2,
    \end{align*}
    \begin{align*}
    & u_h(\human) -u_h(\hm) \leq u_h(\hmonte) - u_h(\hm) +\sigma_1 +  \frac{\sigma_2}{\Phuman(\hm)} ,\\
    & \leq u_h(\hmonte)-u_m(\hmonte) + u_m(\hmonte) -u_m(\hm) + u_m(\hm) - u_h(\hm)  \\
    &+\sigma_1 +  \frac{\sigma_2}{\Phuman(\hm)} ,\\
    &  \leq u_m(\hmonte) -u_m(\hm)+ 2\sqrt{\frac{1}{2|D|} \log \frac{8}{\delta}}  +\sigma_1 +  \frac{\sigma_2}{\Phuman(\hm)}, \\
    &  \leq u_m(\hmonte) -u_m(\hm)+ O\left(\max \left(\frac{1}{\sqrt{n}}, \frac{1}{\sqrt{D}}\right)\right).
\end{align*}

\paragraph{From this, we assume that the MAP decoding target is the true value.}

\begin{align*}
    \Phuman(\hmonte)u_h(\human) - \Phuman(\hmonte)u_h(\hmonte) &\leq \Phuman(\hmonte) \cdot \sigma_1.
\end{align*}
\begin{align*}
    \Phuman(\maphuman)u_h(\hmonte) - \Phuman(\hm)u_h(\hmonte)&\leq u_h(\hmonte)\cdot \sigma_2. 
\end{align*}
Combined above formulation:
\begin{align*}
    \Phuman(\maphuman)u_h(\hmonte) - \Phuman(\hmonte)u_h(\hmonte)&\leq  \Phuman(\hmonte) \sigma_1 + u_h(\hmonte)\sigma_2 + \Phuman(\hm)u_h(\hmonte) \\
    &- \Phuman(\hmonte)u_h(\human)
    \end{align*}
    \begin{align*}
    \Phuman(\maphuman) - \Phuman(\hmonte) &\leq \Phuman(\hm)- \Phuman(\hmonte) \frac{u_h(\human)}{u_h(\hmonte)} + \frac{\Phuman(\hmonte)}{u_h(\hmonte)}\sigma_1 + \sigma_2\\
     &\leq \Phuman(\hm) -\Pmodel(\hm)+\Pmodel(\hm) -\Pmodel(\hmonte) \\
    &+ \Pmodel(\hmonte) - \Phuman(\hmonte)  + \frac{\sigma_1}{u_h(\hmonte)} + \sigma_2\\
 &\leq 4\sqrt{\frac{1}{2|D|}\log \frac{8}{\delta}} + \frac{\sigma_1}{u_h(\hmonte)} + \sigma_2 + \Pmodel(\hm) -\Pmodel(\hmonte)\\
   &\leq O\left(\max \left(\frac{1}{\sqrt{n}}, \frac{1}{\sqrt{D}}\right)\right)+ \Pmodel(\hm) -\Pmodel(\hmonte).\\
\end{align*}
\subsection{Proof of Observation 2}\label{appendix:observation2}
Remid of $\mathrm{Regret}^\mathrm{MAP}_{n,D}$ and $\operatorname{Regret}_{n,D}$ 's upper bound:
\begin{align*}
   \mathrm{Regret}^\mathrm{MAP}_{n,D}&\leq \underbrace{ 8\sqrt{\frac{1}{n}\log\frac{1}{\delta}}+8\sqrt{\frac{1}{|D|}\log\frac{1}{\delta}}}_{\phi_1}.
   \end{align*}

\begin{align*}
& \operatorname{Regret}_{n,D}  \leq  \underbrace{4 \sqrt{\frac{1}{n} \log \frac{1}{\delta}}+4\sqrt{\frac{1}{|D|} \log \frac{1}{\delta}} 
        + \frac{36 }{n}\sqrt{d \log  d}}_{\phi_2}.
\end{align*}

\begin{align*}
    &\phi_1- \phi_2 = 4 \sqrt{\frac{1}{n} \log \frac{1}{\delta}}+4\sqrt{\frac{1}{|D|} \log \frac{1}{\delta}} 
        - \frac{36 }{n}\sqrt{d \log  d}.
\end{align*}

From the above inequality, we can derive this observation.
\begin{itemize}
    \item $n\rightarrow \infty$ and $D$ is finite, $\mathrm{Regret}_{n,D}$ < $\mathrm{Regret}^\mathrm{MAP}_{n,D}$.
    \item $D$ is infinite, $n$ is finite, $\mathrm{Regret}_{n,D}$ < $\mathrm{Regret}^\mathrm{MAP}_{n,D}$.
    \begin{align*}
    \frac{1}{9}\sqrt{n\log{\frac{1}{\delta}}} \geq \sqrt{d \log d}.
\end{align*}
\item $D$ is finite and $n$ is finite, 
the upper bound of $\mathrm{Regret}_{n,D}$ remains less than or equal to the upper bound of $\mathrm{Regret}^\mathrm{MAP}_{n,D}$, provided the following condition holds:
\begin{align*}
    \frac{n}{9}\left(\sqrt{\frac{1}{n} \log \frac{1}{\delta}}+\sqrt{\frac{1}{|D|} \log \frac{1}{\delta}}\right) &\geq \sqrt{d \log d}.
\end{align*}
\end{itemize}

\section{If the utility function can be considered as a kernel function.}\label{appendix:kernel}



Using the Assumption~\ref {assumption:kernel}, we can get the following corollary related to the upper bound of Rademacher complexity $\mathcal{R}_n(\mathcal{F})$.

\begin{maincoro}
    Under the Assumption~\ref {assumption:kernel}, the upper bound of Rademacher complexity $\mathcal{R}_n(\mathcal{F})$ follow as:
    \begin{align}
        \mathcal{R}_n(\mathcal{F}) &\leq  \frac{1}{\sqrt{n}}.\label{eq:7}
    \end{align}
\end{maincoro}

\begin{proof}
We start by finding the Rademacher complexity $\mathcal{R}_n(\mathcal{F})$.
\begin{align}
\mathcal{R}_n(\mathcal{F}) &= \E\left[\max_{h \in \mathcal{F}} \frac{| \sum_{i=1}^n  \epsilon_i u(h,y_i)|}{n}\right]\label{eq:1} \\
& = \E\left[\max_{h \in \mathcal{F}} \frac{  |\left\langle\sum_{i=1}^n \epsilon_i u\left(y_i, \cdot\right), u(h,\cdot)\right\rangle_{\mathcal{F}}|}{n}\right]\label{eq:2} \\
& \leq \E\left[\max_{h \in \mathcal{F}} \frac{\left\|\sum_{i=1}^n \epsilon_i u\left(y_i, \cdot\right)\right\|_{\mathcal{F}}\|u(h,\cdot)\|_{\mathcal{F}}}{n}\right]\label{eq:3} \\
&\leq \E\left[\frac{\sqrt{\Umax}\left\|\sum_{i=1}^n \epsilon_i u\left(y_i, \cdot\right)\right\|_{\mathcal{F}}}{n}\right]\label{eq:4} \\
& =\E\left[\frac{\sqrt{\Umax}\sqrt{\sum_{i, j=1}^n \epsilon_i \epsilon_j u\left(y_i, y_j\right)}}{n}\right]\label{eq:5}\\
&\leq \frac{\sqrt{\Umax}\sqrt{\E\left[\sum_{i, j=1}^n \epsilon_i \epsilon_j u\left(y_i, y_j\right)\right]}}{n}\label{eq:6} \\
& =\frac{\sqrt{\Umax}\sqrt{\sum_{i=1}^n u\left(y_i, y_i\right)}}{n} \leq \frac{\Umax}{\sqrt{n}}.\label{eq:7}
\end{align}

\begin{itemize}
    \item Eq.~(\ref{eq:1}) to Eq.~(\ref{eq:2}) uses the reproducing property $u\left(h, y_i\right)=\left\langle u\left(y_i, \cdot\right), u(h, \cdot)\right\rangle_{\mathcal{F}}$.
    \item Eq.~(\ref{eq:2}) to Eq.~(\ref{eq:3}) uses the Cauchy Schwarz inequality $|\langle \sum_{i=1}^n \epsilon_i u\left(y_i, \cdot\right), u(h,\cdot)\rangle|_{\mathcal{F}} \leq\|\sum_{i=1}^n \epsilon_i u\left(y_i, \cdot\right)\|_{\mathcal{F}}\|u(h,\cdot)\|_{\mathcal{F}}$
    \item Eq.~(\ref{eq:3}) to Eq.~(\ref{eq:4}) uses $\|u(h,\cdot)\|_{\mathcal{F}} = \sqrt{u(h,h)} \leq \sqrt{\Umax}$ 
    \item  Eq.~(\ref{eq:4}) to Eq.~(\ref{eq:5}) uses $\left\|\sum_{i=1}^n \epsilon_i u\left(y_i, \cdot\right)\right\|_{\mathcal{F}}\overset{\text{reproducing property}}{=}\sqrt{\sum_{i, j=1}^n \epsilon_i \epsilon_j\left\langle u\left(y_i, \cdot\right), u\left(y_j, \cdot\right)\right\rangle_{\mathcal{F}}} = \sqrt{\sum_{i, j=1}^n \epsilon_i \epsilon_j u\left(y_i, y_j\right)}$
    \item Eq.~(\ref{eq:5}) to Eq.~(\ref{eq:6}) uses Jensen Inequality.
    \item Eq.~(\ref{eq:6}) to Eq.~(\ref{eq:7}), $\E\left[\sum_{i, j=1}^n \epsilon_i \epsilon_j u\left(y_i, y_j\right)\right]=\sum_{i, j=1}^n \E\left[\epsilon_i \epsilon_j\right] u\left(y_i, y_j\right) = \sum_{i=1}^n u\left(y_i, y_i\right)$
    
\end{itemize}
\end{proof}

We know the following bound from Appendix~\ref{apendix:heart}.

\begin{align*}
    \Delta(u_m,\widehat{u}_m,\human)+\Delta(\widehat{u}_m,u_m,\hmonte)& \leq2 \mathcal{R}_n(\mathcal{F}) +2\Umax\sqrt{\frac{1}{2n}\log \left(\frac{4}{\delta}\right)}.\\
    &\leq 3\sqrt{\frac{1}{n} \log \frac{1}{\delta}} +\frac{2}{\sqrt{n}}.
\end{align*}

\section{Experimental Details of Numerical Simulation (Section~\ref{sec:experiments})}\label{appendix:exp}
$\Phuman$ is non-uniform distribution (reflecting real-world biases), for each seed, we generate $\Phuman$ via i.i.d. sampling, then form the empirical model distribution $\Pmodel$ by drawing $D$ times from $\Phuman$ (i.e, $\widehat{P}$ represents hypothesis frequencies).
In the experiment setting of Fig.~\ref{fig:wasserstein_distance}, we applied the utility function according to Assumption~\ref{assumption:utility}, and in  Fig.~\ref{fig:fix_d} and Fig.~\ref{fig:fix_n}, we assume a symmetric utility matrix $u\in\mathbb{R}^{\mathcal{Y}\times\mathcal{Y}}$ with $u(i,i)=1$ and $u(i,j)\in[0,1]$ for $i\neq j$, assigning slightly higher utilities to outcomes with higher $\Phuman$ probabilities.

\end{document}